\documentclass{iopconfser}
\usepackage{amsmath}
\usepackage{graphicx}
\usepackage{subcaption}
\usepackage{hyperref}
\newcommand\blfootnote[1]{%
  \begingroup
  \renewcommand\thefootnote{}\footnote{#1}%
  \addtocounter{footnote}{-1}%
  \endgroup
}

\begin{document}

\title{Computer Vision-Based Early Detection of Container Loss at Sea  \blfootnote{Thanks to Singapore Maritime Institute (SMI) for supporting this work.}}

\author{Vishakha Lall$^{1}$, Capt. Stanley S Pinto$^{2}$, Capt. Chu Xing Peng$^{2}$, Wu Kaiwen$^{1}$}

\affil{$^1$Centre of Excellence in Maritime Safety, Singapore Polytechnic, Singapore}

\affil{$^2$Quality, Safety, Security and Environment, Pacific International Lines (Pvt) Ltd, Singapore}

\email{vishakha\_lall@sp.edu.sg}

\begin{abstract}
Containerised shipping underpins global trade, yet container loss at sea remains a persistent safety, environmental, and economic challenge. Despite compliance with Cargo Securing Manuals, dynamic maritime conditions such as vessel motion, wind loading, and severe sea states can progressively destabilise container stacks, leading to overboard losses. With the new International Maritime Organisation's (IMO) mandatory reporting requirements for lost containers, there is an urgent need for a reliable, evidence-based early detection solution for destabilised containers. This study showcases a low-cost, retrofittable computer vision–based system for early detection of destabilised containers using existing onboard cameras. The framework integrates object segmentation to isolate container stacks, temporal object tracking using optical flow and individual objects' residual motion extraction to quantify relative movement. Experimental evaluation on real onboard ship footage demonstrates that the proposed pipeline effectively isolates container-level motion under challenging conditions of varying sea states and visibility conditions. By enabling early alerts for crew intervention and navigational adjustment, the proposed approach enhances cargo safety, operational resilience, and regulatory compliance.
\end{abstract}

\section{Introduction}

Container loss at sea is principally driven by two interrelated factors: deficiencies in cargo securing practices as described in the vessel’s Cargo Securing Manual (CSM) and the dynamic environmental loads encountered during a voyage \cite{container_loss_causes}. The CSM sets out mandatory requirements for container stack configuration, permissible stack weights, and the application of securing devices such as twist locks and lashing systems, and compliance with these provisions can be effectively regulated and enforced through established port, class, and shipboard procedures. In contrast, the environmental conditions encountered at sea, characterised by strong winds, large waves, and dynamic phenomena such as parametric and synchronous rolling, are inherently beyond operational control and can impose dynamic loads on container stacks that exceed the assumptions underpinning CSM design limits \cite{forces_on_containers}. Empirical analyses of incident data between 2011 and 2023 reveal that adverse weather conditions are the predominant cause of container loss, accounting for 57.14\% of incidents \cite{weather_as_reason_for_container_loss}, underscoring that even with effective enforcement of the CSM, environmental forcing remains an inherent and intensifying risk factor in global container ship operations.  

The loss of containers at sea has wide-ranging economic, safety, regulatory, and environmental consequences. Each container loss event represents a substantial financial burden, encompassing the direct loss of cargo, often valued in the millions of dollars per incident, as well as secondary costs related to insurance claims, salvage operations, voyage delays, and potential liability for damage to third parties. At a systemic level, these losses contribute to increased insurance premiums and operational costs across the global container shipping industry. From a navigational safety perspective, lost containers, particularly those that remain partially submerged, pose a serious collision hazard to other vessels. Such containers are difficult to detect visually and by conventional radar, increasing the risk of hull damage, propulsion failure, or catastrophic accidents \cite{floating_containers}. The vastness of the ocean, coupled with historically incomplete reporting and limited tracking capabilities, further complicates efforts to locate and recover lost containers, allowing many to remain unaccounted for over extended periods. In addition to navigational hazards, container loss presents significant environmental risks. Containers carrying hazardous or polluting cargoes may rupture, releasing chemicals, plastics, or other harmful substances into the marine environment, with long-term consequences for marine ecosystems and coastal regions \cite{environmental_impact_of_lost_containers, environmental_impact_of_lost_containers_2}. Even non-hazardous cargo can contribute to marine debris, exacerbating pollution and posing ingestion and entanglement risks to marine life. These environmental impacts reinforce the need for improved monitoring, reporting, and mitigation strategies to address container loss.

The frequency of container loss, therefore, remains a persistent concern. As per the World Shipping Council's Container Loss Report - 2025 Update, the most notable spike occurred in 2013, when 5,578 containers were lost, by far the highest. Other years with elevated loss figures include 2020 (3,924) and 2021 (2,301), which were periods marked by several large-scale incidents. In contrast, 2023 saw a significant reduction with 221 containers lost. But in 2024, the total rose to 576 containers, highlighting that container loss is not an isolated or rare phenomenon but a recurring operational risk in global maritime transport. Recognising the implications of underreporting, the International Maritime Organisation (IMO) has adopted amendments to the Safety of Life at Sea (SOLAS) Convention, mandating the reporting of container losses from January 2026 onwards.

Existing research on container loss at sea has predominantly focused on post-loss analysis, encompassing incident investigations, environmental impact assessments, and methods for detecting and mitigating hazards posed by containers after they have entered the marine environment. Representative studies, as summarised in \cite{existing_methods}, examine local detection systems, including radar, sonar, and camera-based observation, for identifying nearby floating or partially submerged containers, as well as remote monitoring systems that rely on embedded communication devices to transmit container location and status information. While these approaches enhance collision avoidance, situational awareness, and long-term tracking, they are inherently reactive, operating only after container loss has occurred. Related work by \cite{wireless_detection_of_lost_containers} advances this post-loss paradigm by integrating container-mounted tracking units (CTDs) equipped with sensors and communication modules to detect when a container goes overboard automatically and broadcast alerts to the host vessel, nearby ships, and shore-based authorities via existing maritime communication standards, including mobile and satellite links. Although such systems improve the timeliness and effectiveness of post-incident response, they do not address the conditions leading up to loss events. Recent studies have also explored the application of advanced artificial intelligence techniques for post-loss container detection. In particular, \cite{submerged_container_segmentation} proposes a submerged container detection framework that combines the You Only Look Once (YOLO) object detection model with the Segment Anything Model (SAM) to identify containers in water-column imagery acquired from sonar or related sensing modalities. Experimental results demonstrate high accuracy in container localisation and segmentation, highlighting that modern deep learning and segmentation models can reliably detect containers in complex and cluttered marine environments, including scenarios involving partial submersion. While these results clearly motivate the use of state-of-the-art computer vision models for accurate container detection, the approach remains post-analytical, relying on data collected after containers have already been lost and offering no mechanism for early warning or loss prevention. In contrast to these reactive approaches, limited prior work has investigated pre-loss monitoring of container stability during voyage. An early example is presented in \cite{vibration_sensor}, which explores the use of wireless vibration sensor tags affixed to containers to continuously monitor dynamic responses associated with stability degradation, such as excessive sway or tipping that may precede cargo shifts or stack destabilisation. While this study demonstrates the feasibility of using wireless sensing to provide early indicators of potential stability loss, it does not address challenges related to large-scale deployment, integration with shipboard systems, or operational scalability.

Based on the insights from domain experts from Pacific Interational Lines (PIL), early notification of container instability is consistently identified as a critical enabler of effective corrective action. Experts emphasised that the time window between the onset of container instability and actual loss overboard can be extremely short, on the order of 60–90 seconds in worst-case scenarios, yet this window is often sufficient for meaningful intervention if reliable alerts are available. Early detection allows the bridge team to implement relatively small but high-impact measures, such as minor course or speed adjustments, which have been shown to significantly reduce vessel roll amplitudes and associated dynamic loads on container stacks. In conditions where weather permits, timely alerts also enable onboard crew to physically inspect and retighten container lashings before progressive failure occurs. Even when corrective actions are insufficient to prevent loss, early detection provides precise timestamps that can support more accurate estimation of loss locations for post-incident response and recovery. Experts further highlighted that such continuous, automated monitoring is particularly valuable during low-visibility or nighttime operations, where manual observation from the bridge may fail to identify early signs of instability. Overall, these discussions reinforce that shifting container monitoring from post-loss analysis to proactive, real-time instability detection can materially reduce risk, enhance crew situational awareness, and improve both safety and operational decision-making.

This paper presents an integrated vision-based framework for detecting destabilised shipping containers using onboard monocular video. The proposed approach combines instance segmentation, multi-object tracking, global motion compensation, and dense optical flow analysis to distinguish true container motion from vessel-induced camera motion. A YOLO-based segmentation model is used to extract precise container masks, which are temporally associated across frames using DeepSORT tracking. To mitigate dominant camera motion caused by ship dynamics, an affine global motion compensation module estimates and removes background motion while explicitly excluding segmented container regions. Dense optical flow is then computed between compensated frames, and displacement is measured within each container mask to capture physically meaningful lateral container movement. A robust statistical formulation is applied in which common motion across all tracked containers is removed using a median-based estimator, yielding per-container residual motion signals. These residuals are accumulated over time and filtered using an interquartile range (IQR)–based outlier suppression scheme to improve robustness against noise and transient flow artifacts. Containers are classified as destabilised based on sustained residual motion exceeding adaptive thresholds over temporal windows. 

\section{Dataset}
The dataset used in this study comprises 100 videos of duration 1 minute each collected from historical onboard video feeds provided by 11 vessels operated by Pacific International Lines (PIL). The videos were captured using fixed cameras installed on the bridge wing port and bridge wing starboard positions, which provide unobstructed lateral views of container stacks along the vessel’s deck. The collected videos span a wide range of operational and environmental conditions to reflect real-world deployment scenarios. Recordings include daytime and nighttime operations, varying lighting conditions, rainfall, reduced visibility, and diverse sea states. This variability introduces significant challenges such as motion blur, low contrast, reflections, and camera vibrations induced by vessel motion, making the dataset representative of practical maritime environments.

As container loss incidents are rare and safety-critical events, no videos in the dataset contain full container loss occurrences. However, through manual inspection of the historical footage, 38 videos were identified in which small but observable container movements are present. These movements correspond to subtle lateral shifts or oscillations of containers that may precede more severe instability. The detection of such early-stage motion forms the primary target of the proposed system, with the objective of generating timely alerts.

This dataset, therefore, reflects a highly imbalanced but realistic operational setting, where the vast majority of video data contains stable container stacks and only a small fraction exhibits detectable motion.

\subsection{Segmentation Model Training Dataset}
\label{segmentation_model_training_dataset}
Given that the majority of the collected videos exhibit minimal observable container motion, consecutive frames are often visually identical or highly redundant. To avoid biasing the segmentation model with near-duplicate samples and to ensure sufficient visual diversity for effective training, a frame sampling strategy was adopted. Approximately 200 frames were extracted from the video footage of each vessel, with sampling performed across different temporal segments to capture variations in viewpoint, illumination, and background conditions. The selected frames were manually annotated at the instance level by delineating precise YOLO-style polygonal segmentation masks for individual containers. This process resulted in a curated dataset of approximately 2,200 annotated frames. 

\subsection{Baseline Data for Model Comparison}
\label{simulated_test_dataset}
In the absence of publicly available benchmark datasets for container instability detection, we evaluate our proposed method using simulated video publicly released by Eyesea and EVI Safety Technologies. This simulated video depicts a container loss event and has been used to demonstrate post-event detection capabilities. While the referenced video focuses on identifying containers after they have fallen, we apply our framework to the same video with a different objective: early-stage instability detection. Specifically, we analyse the temporal segments preceding the container loss events to assess the model’s ability to identify subtle container movements indicative of instability before visible collapse occurs. This enables a qualitative comparison between post-loss detection approaches and our proposed early-warning methodology.

\section{Proposed Methodology}

\subsection{Container Segmentation}
To detect early-stage container instability, it is necessary to identify and track the motion of individual containers over time. In operational settings, slackened lashings typically result in subtle, localised movement affecting single containers rather than entire stacks. For this reason, an instance segmentation approach is adopted. Conventional object detection methods localise containers using bounding boxes, which remain largely unchanged under small intra-object movements. While effective for detecting fallen or falling containers, such representations lack the sensitivity required for early instability prediction. Instance segmentation overcomes this limitation by providing pixel-level masks that precisely delineate each container’s shape, enabling motion analysis to be confined to container-specific regions. This mask-level representation enables the downstream detection of subtle displacements even when bounding boxes remain static.

In this work, we employ the YOLO11m-seg model \cite{yolo11_ultralytics}, which comprises approximately 22B parameters and offers a favourable trade-off between segmentation accuracy and inference latency, making it suitable for real-time deployment. The model is fine-tuned on the curated container segmentation dataset (Sec. \ref{segmentation_model_training_dataset}) using an $80:10:10$ training–validation-test split. Training is conducted on an NVIDIA GeForce RTX 4070 GPU with a batch size of $40$ for $200$ epochs, using a learning rate of $0.0005$ and momentum of $0.9$. All input frames are resized to $400 \times 400$ pixels. During inference, batch processing is employed with a batch size of $200$ to maximise throughput.

\subsection{Temporal Container Tracking}
Accurate assessment of container stability requires consistent identification of individual containers across time. Given the dense arrangement of container stacks, frequent partial occlusions, and minimal inter-frame appearance changes, simple frame-wise detection is insufficient. We therefore employ a temporal tracking framework to maintain container identities across frames and enable motion aggregation over time.

From each segmentation result, an axis-aligned bounding box is derived and used as the spatial representation required by the tracker. These detections are then passed to DeepSORT \cite{deepsort_1, deepsort_2}, which performs online multi-object tracking by associating detections across consecutive frames. DeepSORT combines motion-based prediction with appearance-based association. A Kalman filter is used to predict the short-term motion of each container, while a cosine-distance metric over learned appearance embeddings is used to resolve ambiguous associations, particularly in scenarios with overlapping or visually similar containers. Tracks are confirmed only after consistent observations across multiple frames, reducing spurious identities caused by transient detections.

To associate segmentation masks with persistent track identities, each confirmed track is matched to the corresponding segmentation instance using Intersection-over-Union (IoU) between the track’s predicted bounding box and the bounding boxes enclosing segmentation masks. This mapping enables the propagation of a unique track identifier to the corresponding segmentation mask, allowing container-specific measurements to be accumulated over time.

\subsection{Global Affine Motion Compensation}
Ship-mounted cameras experience continuous ego-motion induced by vessel dynamics, including roll, pitch, yaw, and translational drift. These global motions introduce apparent movement across the entire image plane, which can obscure the subtle relative motion of individual containers. Since early-stage container destabilisation manifests as small, local displacements relative to neighbouring containers rather than large absolute motion, it is necessary to explicitly remove global camera-induced motion before analysing object-level dynamics. To address this, we apply global affine motion compensation (GMC) between consecutive frames to estimate and subtract camera motion, thereby isolating residual motion attributable to individual containers.

Let $I_{t-1}$ and $I_t$ denote two consecutive video frames. Global camera motion between these frames is modelled using a 2D affine transformation:
\begin{equation}
\label{global_affine_transformation}
    \begin{bmatrix}
    x_t\\
    y_t
    \end{bmatrix}
    = \begin{bmatrix}
    a_{00} & a_{01}\\
    a_{10} & a_{11}
    \end{bmatrix}
    \begin{bmatrix}
    x_{t-1}\\
    y_{t-1}
    \end{bmatrix}
    +\begin{bmatrix}
    b_x\\
    b_y
    \end{bmatrix}
\end{equation}
where $(x_t,y_t)$ and $(x_{t-1},y_{t-1})$ are corresponding pixel locations in $I_t$ and $I_{t-1}$ respectively, $(a_{00},a_{01},a_{10},a_{11})$ are affine parameters that capture rotation, scaling, and shear, and $(b_x, b_y)$ represents translation.

Direct estimation of global motion from the full image may be biased by moving containers. To mitigate this, segmentation masks obtained from the container instance segmentation stage are used to construct an exclusion mask, preventing feature extraction and motion estimation within container regions. Let $Bg$ denote the set of pixels not belonging to any container mask. Feature points are extracted only from $Bg$, and matched between $I_{t-1}$ and $I_t$. The affine transformation $M_t$ is estimated using a RANSAC (Random Sample Consensus) estimator to minimise reprojection error:
\begin{equation}
\label{ransac_estimator}
    M_t = arg\,min \sum_{(p,q)\in C} ||q-Mp||^2
\end{equation}
where $(p,q)$ are the corresponding matching features in $I_{t-1}$ and $I_t$ respectively and $C$ denotes the set of matched features among $Bg$. 

The estimated affine transform is applied to warp the current frame:
\begin{equation}
\label{warp_affine}
    \widetilde{I_t} = warp\,affine(I_t,M_t)
\end{equation}
resulting in a motion-compensated frame $\widetilde{I_t}$ that is aligned with $I_{t-1}$ in the camera reference frame.

\subsection{Relative Container Movement}
Following global affine motion compensation, the remaining pixel-level motion primarily reflects object-level dynamics rather than camera-induced movement. To quantify the motion of individual containers, we compute relative container motion using dense optical flow constrained to instance segmentation masks.

Let $\widetilde{I_{t-1}}$ and $\widetilde{I_t}$ denote two consecutive motion-compensated frames. Dense optical flow $F_t$ is computed between these frames using the Farnebäck method \cite{farneback}. 
\begin{equation}
\label{farneback}
    F_t(x,y) = (u_t(x,y), v_t(x,y))
\end{equation}
where $u_t, v_t$ represent the horizontal and vertical pixel displacements at location (x,y).

For each detected container instance $i$, let $S_i$ denote its segmentation mask. The container's absolute motion magnitude is estimated by aggregating flow vectors within the mask:
\begin{equation}
\label{absolute_motion}
    v_i^{abs}(t) = \frac{1}{|S_i|} \sum_{(x,y)\in S_i} u_t(x,y)
\end{equation}
where only the horizontal component is considered, as lateral motion is the dominant indicator of container destabilisation in the observed camera geometry.

After global motion compensation, residual motion may remain due to elastic deformation of container stacks under wave loading and to synchronous sway of multiple containers caused by vessel roll and pitch. To normalise this, we compute a robust estimate of the common motion across all containers to eliminate stack-level or vessel-induced motion that affects all containers simultaneously:
\begin{equation}
\label{common_motion}
    v^{common}(t) = median(\{v_i^{abs}(t)\}_{i=1}^N)
\end{equation}
where $N$ is the number of tracked containers in the frame. The median is used as a robust estimator that is resilient to outlier containers exhibiting abnormal motion.

The relative motion of each container is then computed as:
\begin{equation}
\label{relative_motion}
    v_i^{rel}(t) = v_i^{abs}(t) - v^{common}(t)
\end{equation}

\section{Results}
\subsection{Container Segmentation Results}
\begin{figure}
    \centering
    \includegraphics[width=1\linewidth]{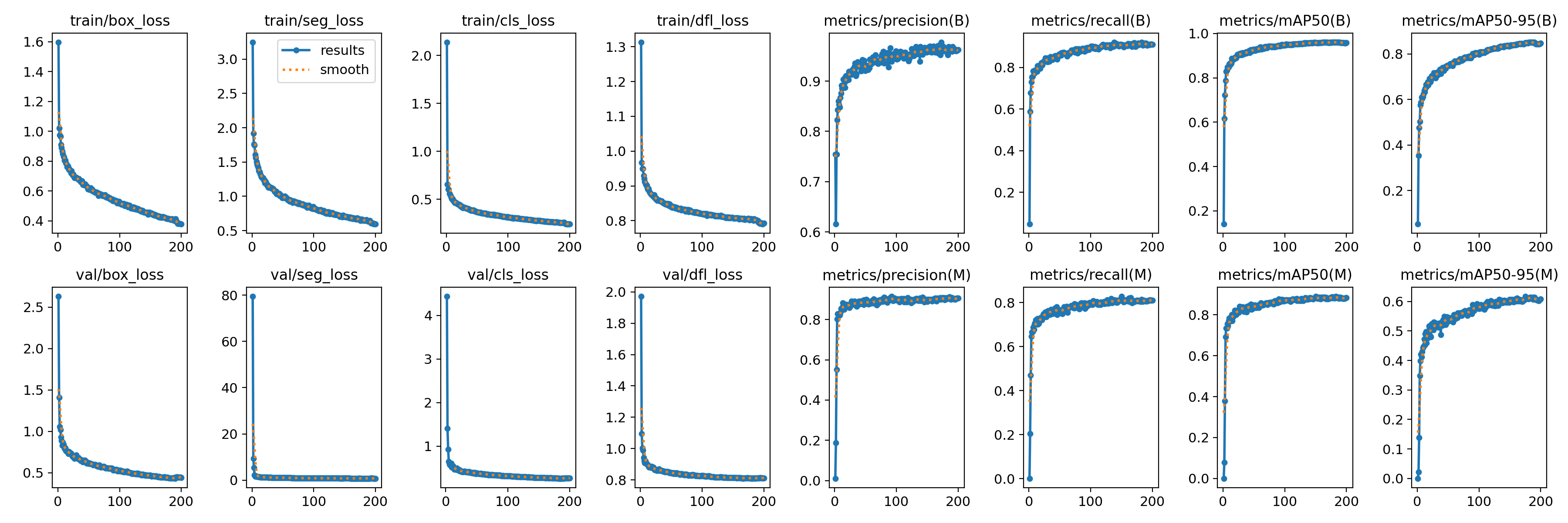}
    \caption{Loss and performance metrics for fine-tuned container segmentation model}
    \label{fig:segmentation_metrics}
\end{figure}

The performance of the fine-tuned container segmentation model is evaluated using standard YOLO segmentation metrics, as shown in Fig. \ref{fig:segmentation_metrics}. Among the reported loss components, bounding box loss and segmentation loss are the most relevant for this work. Accurate bounding box regression is essential for reliable temporal tracking and track–mask association, while high-quality segmentation masks are critical for capturing subtle, container-level motion used in early instability detection. Since the task involves a single object class (containers), classification loss is not a primary concern. Similarly, distribution focal loss (DFL) is less relevant in this context, as the proposed system does not aim to track distant containers where fine-grained localisation uncertainty would dominate. Both bounding box and segmentation losses exhibit a smooth and consistent decrease over training epochs for the training and validation sets, indicating stable convergence without overfitting.

In terms of performance metrics, precision, recall, and mean Average Precision (mAP) are the most relevant. High precision reduces false container detections that could propagate into the tracking and motion analysis stages, while high recall ensures that all visible containers are consistently captured across frames. The model achieves strong performance on mAP@50, demonstrating robust container localisation and segmentation. As expected, mAP@50–95 yields slightly lower values due to stricter overlap thresholds but remains consistently high for both training and validation, confirming the model’s ability to produce accurate instance masks suitable for downstream temporal tracking and motion estimation.

\begin{figure}
  \centering
  \begin{subfigure}{0.22\linewidth}
    \includegraphics[width= \linewidth]{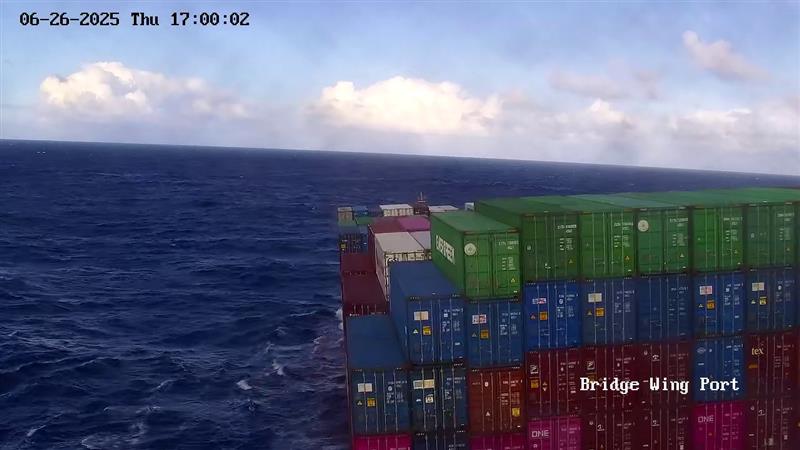}\hfill
    \caption{}
  \end{subfigure}
  \begin{subfigure}{0.22\linewidth}
    \includegraphics[width= \linewidth]{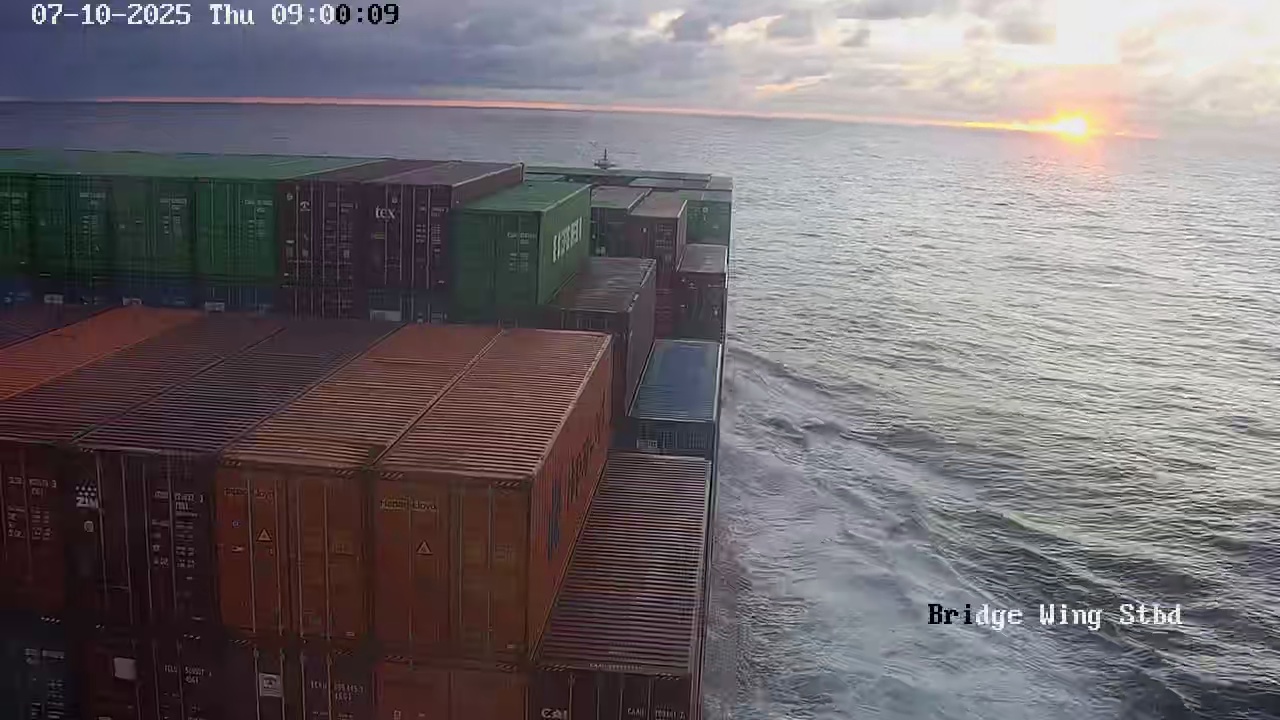}\hfill
    \caption{}
  \end{subfigure}
  \begin{subfigure}{0.22\linewidth}
    \includegraphics[width= \linewidth]{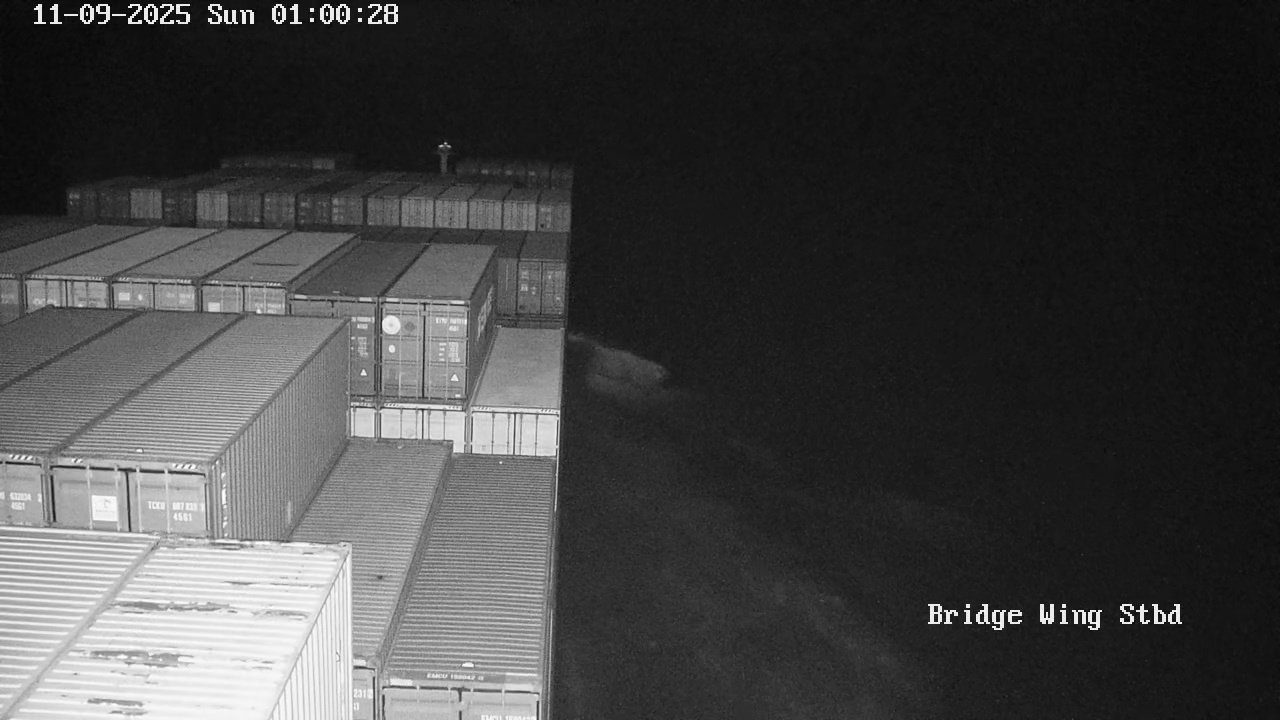}\hfill
    \caption{}
  \end{subfigure}
  \begin{subfigure}{0.22\linewidth}
    \includegraphics[width= \linewidth]{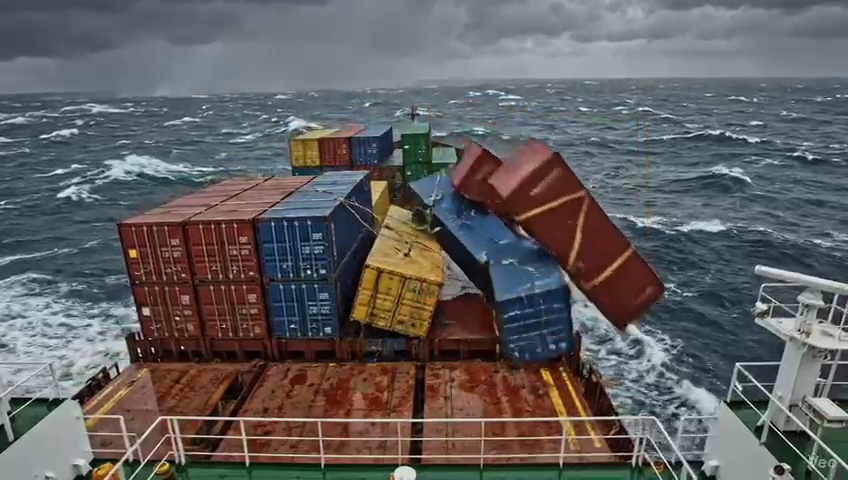}\hfill
    \caption{}
  \end{subfigure}
  \begin{subfigure}{0.22\linewidth}
    \includegraphics[width= \linewidth]{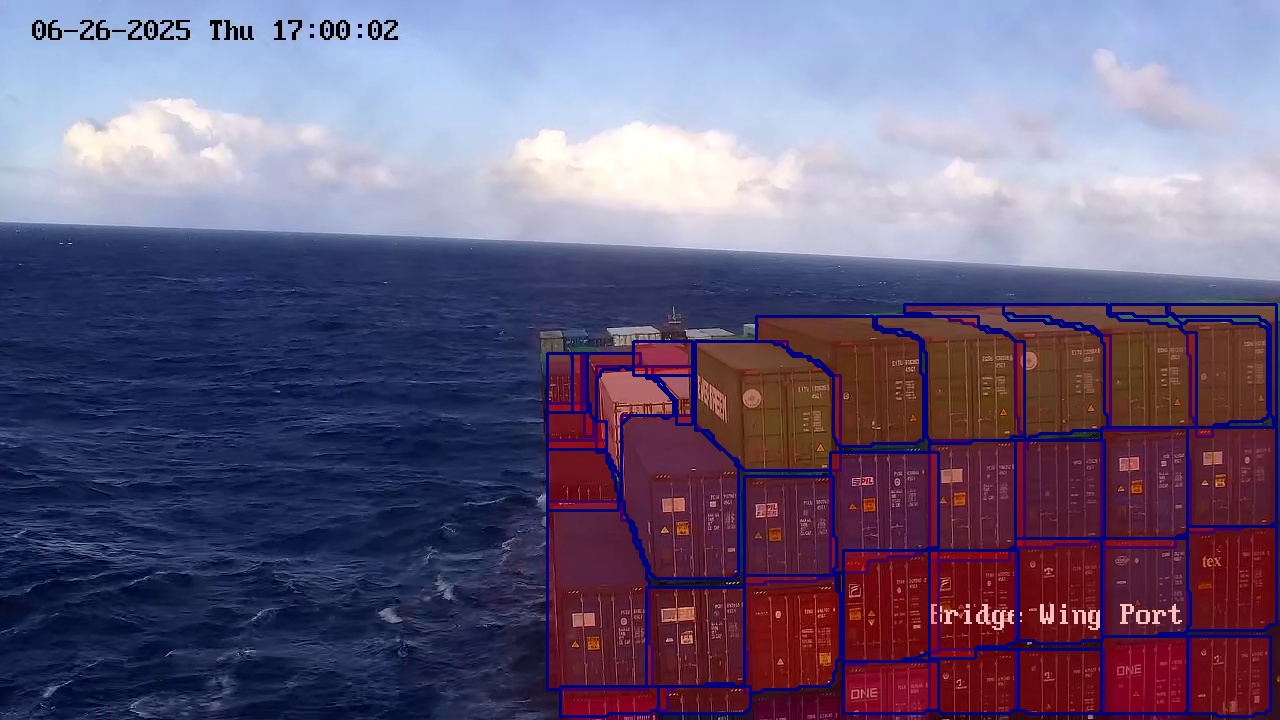}\hfill
    \caption{}
  \end{subfigure}
  \begin{subfigure}{0.22\linewidth}
    \includegraphics[width= \linewidth]{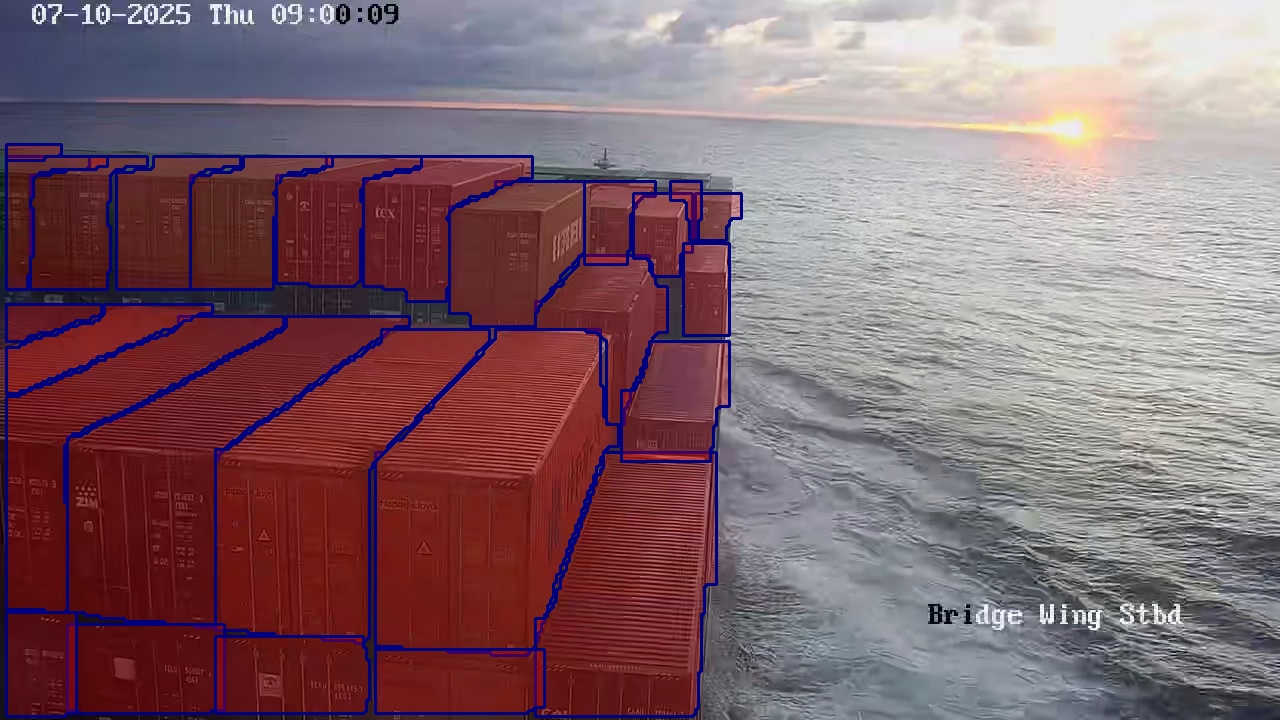}\hfill
    \caption{}
  \end{subfigure}
  \begin{subfigure}{0.22\linewidth}
    \includegraphics[width= \linewidth]{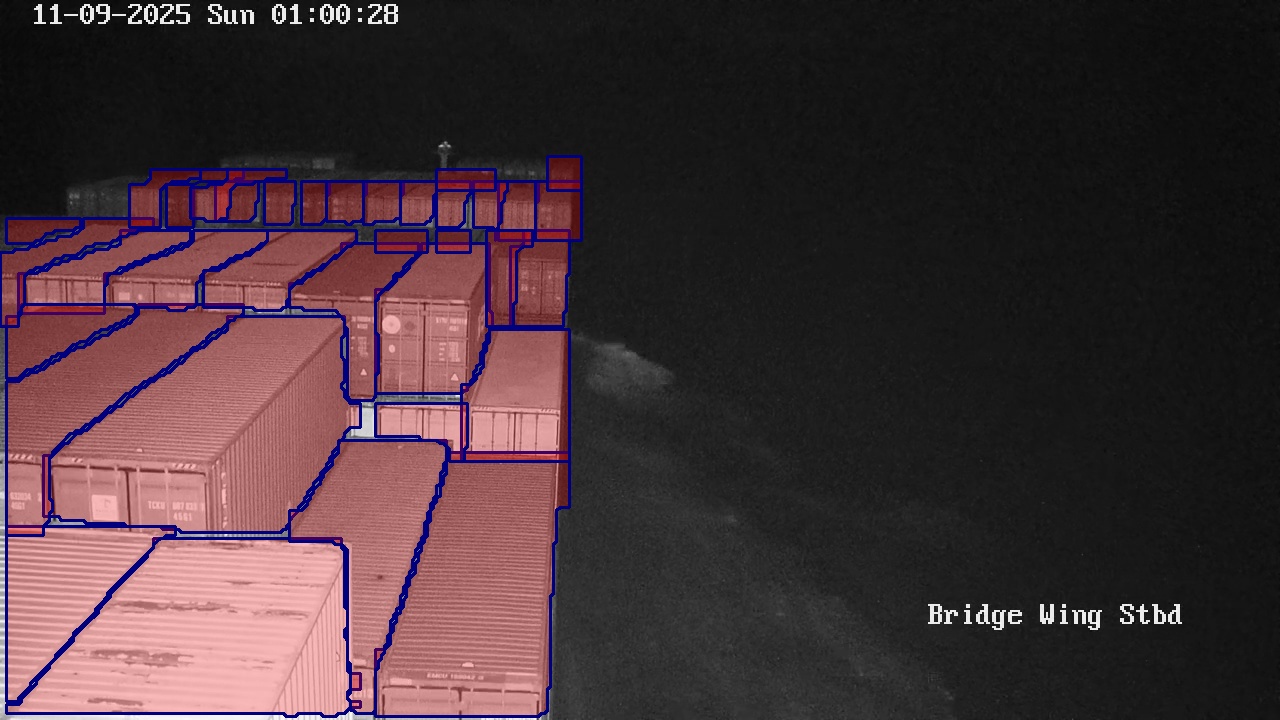}\hfill
    \caption{}
  \end{subfigure}
  \begin{subfigure}{0.22\linewidth}
    \includegraphics[width= \linewidth]{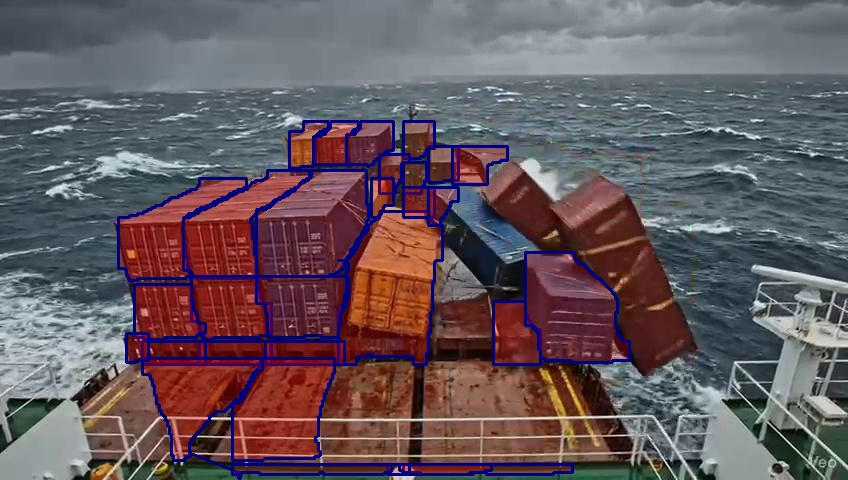}\hfill
    \caption{}
  \end{subfigure}
  \caption{Segmentation samples from test dataset (a,e) original frame and segmented containers in normal lighting, (b,f) original frame and segmented containers in bright lighting, (c,g) original frame and segmented containers in night conditions, all from collected test dataset (Sec. \ref{segmentation_model_training_dataset}), (d,h) original frame and segmented containers from simulated test video (Sec. \ref{simulated_test_dataset})}
  \label{fig:segmentation_samples}
\end{figure}

Fig. \ref{fig:segmentation_samples} presents representative samples from the test dataset, showing the original frames and their corresponding segmentation outputs. The results demonstrate that the model robustly segments individual containers across different camera configurations (bridge wing port and starboard views) and under varying lighting conditions, including day and night operations.

\subsection{Temporal Container Tracking Results}

\begin{figure}
  \centering
  \begin{subfigure}{0.18\linewidth}
    \includegraphics[width= \linewidth]{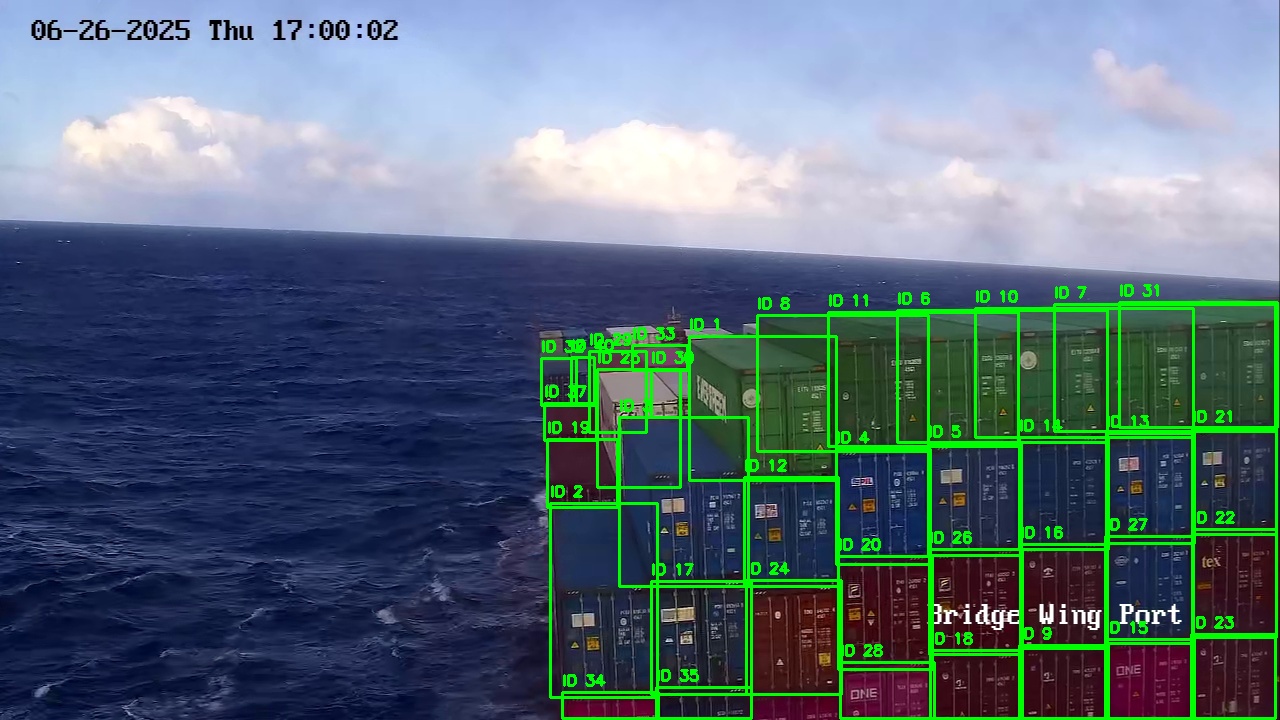}\hfill
    \caption{}
  \end{subfigure}
  \begin{subfigure}{0.18\linewidth}
    \includegraphics[width= \linewidth]{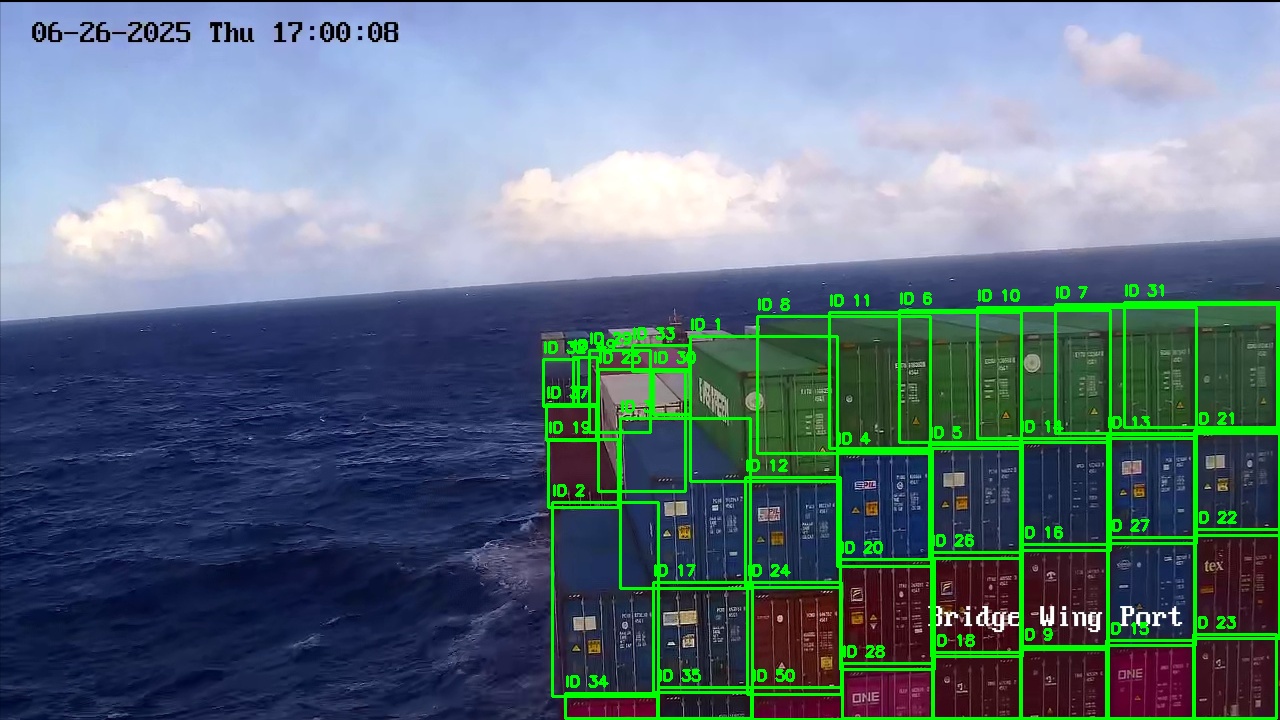}\hfill
    \caption{}
  \end{subfigure}
  \begin{subfigure}{0.18\linewidth}
    \includegraphics[width= \linewidth]{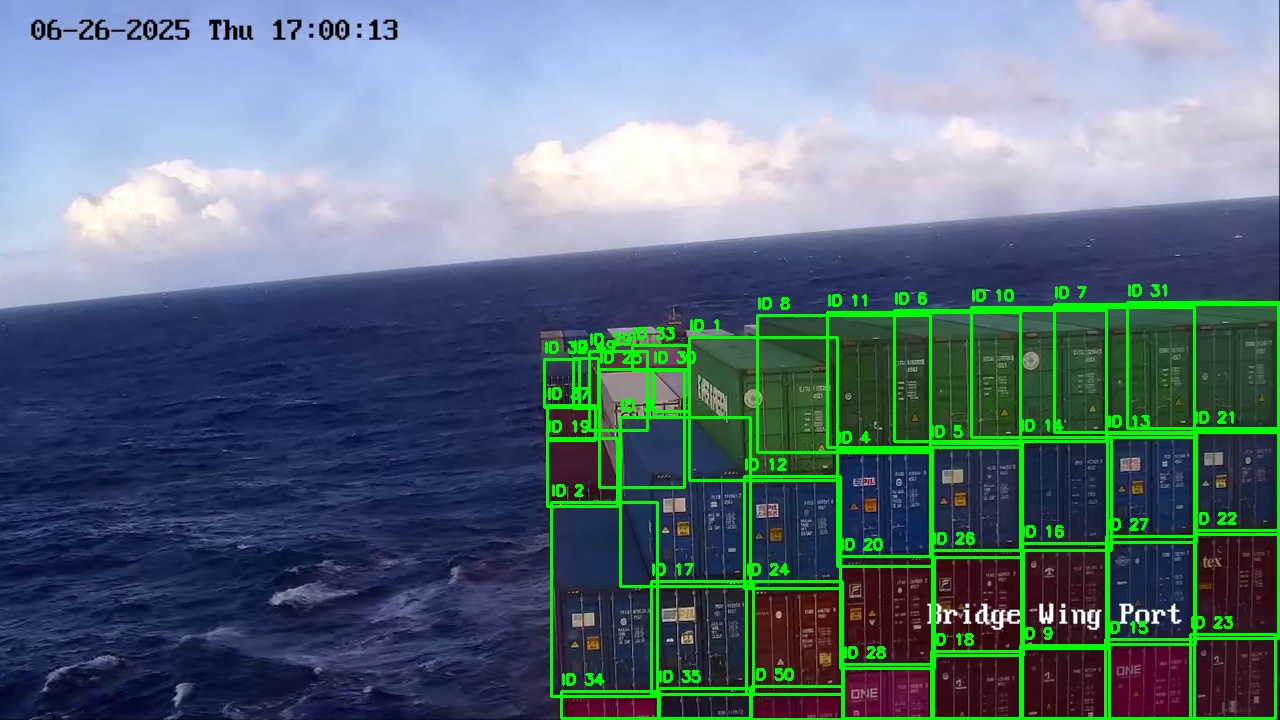}\hfill
    \caption{}
  \end{subfigure}
  \begin{subfigure}{0.18\linewidth}
    \includegraphics[width= \linewidth]{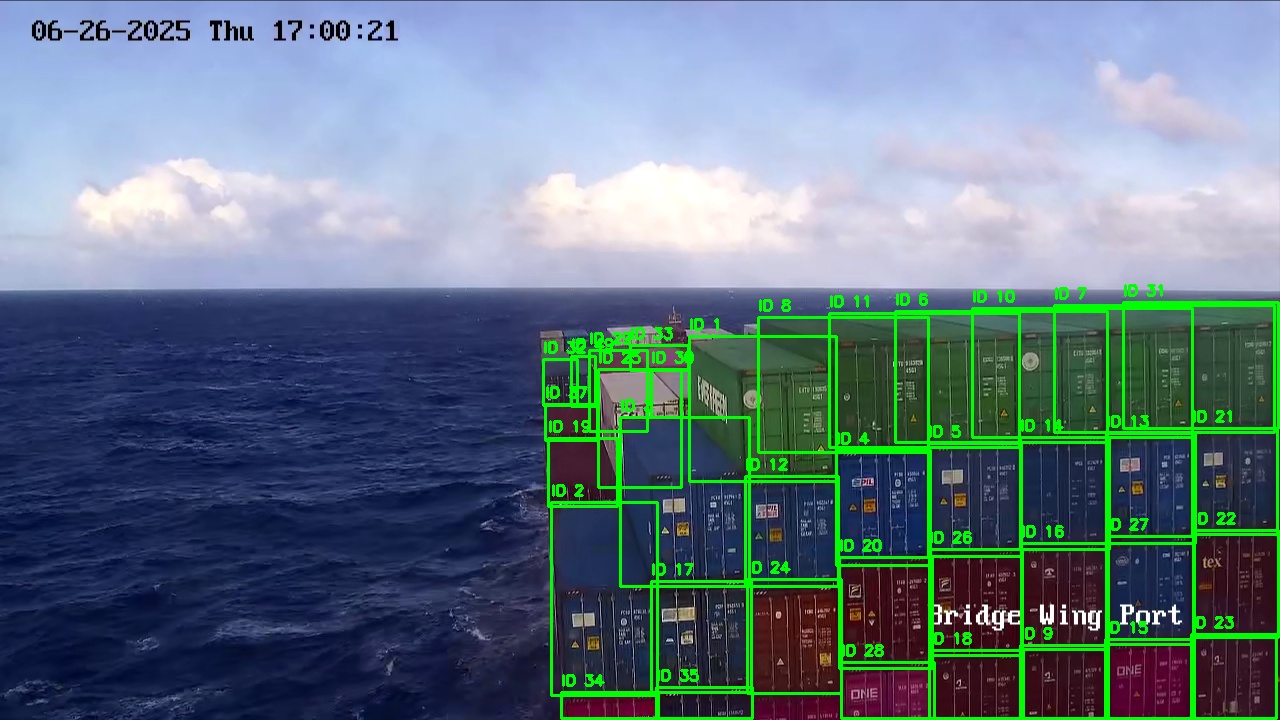}\hfill
    \caption{}
  \end{subfigure}
  \begin{subfigure}{0.18\linewidth}
    \includegraphics[width= \linewidth]{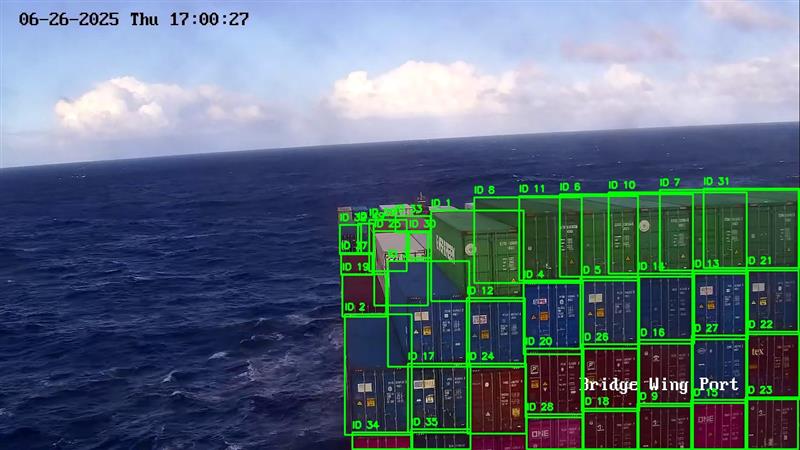}\hfill
    \caption{}
  \end{subfigure}
  \begin{subfigure}{0.18\linewidth}
    \includegraphics[width= \linewidth]{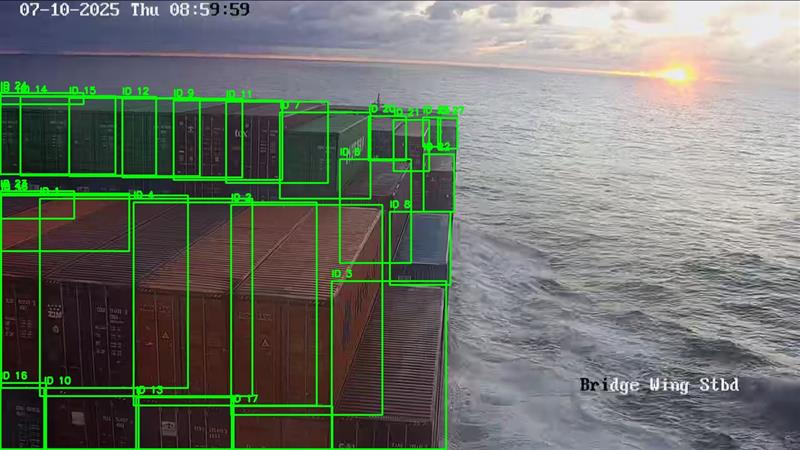}\hfill
    \caption{}
  \end{subfigure}
  \begin{subfigure}{0.18\linewidth}
    \includegraphics[width= \linewidth]{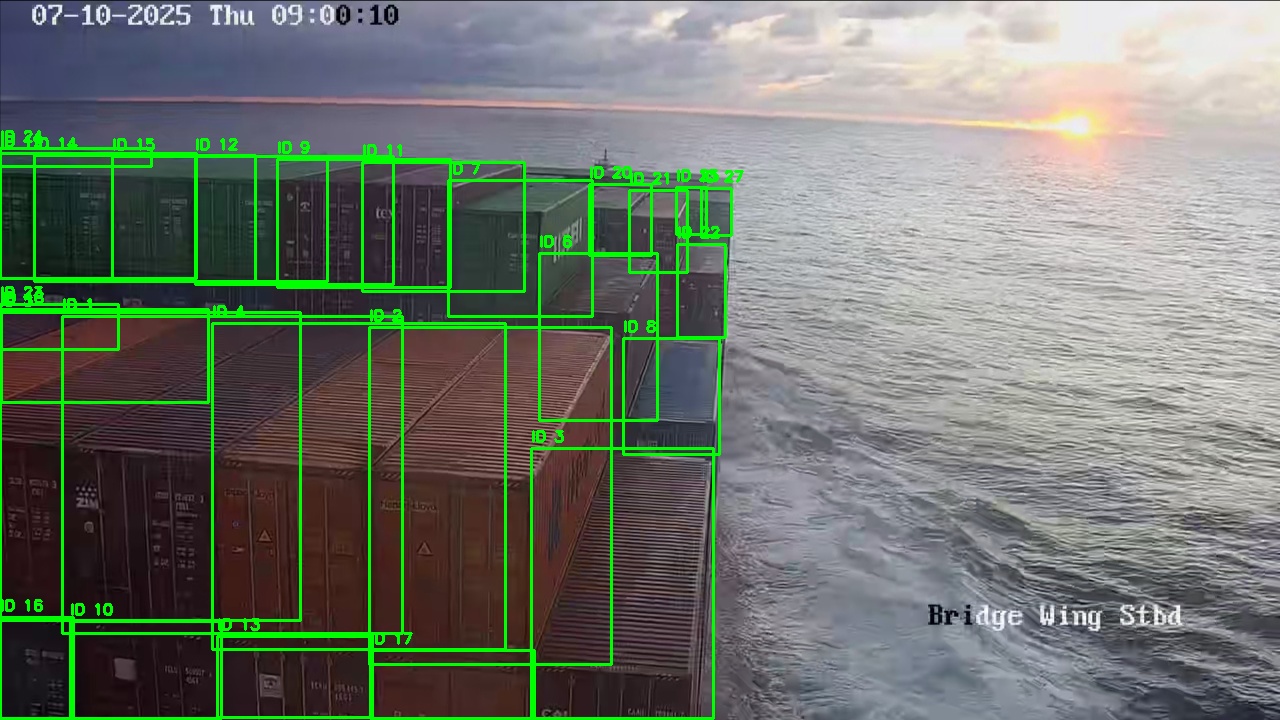}\hfill
    \caption{}
  \end{subfigure}
  \begin{subfigure}{0.18\linewidth}
    \includegraphics[width= \linewidth]{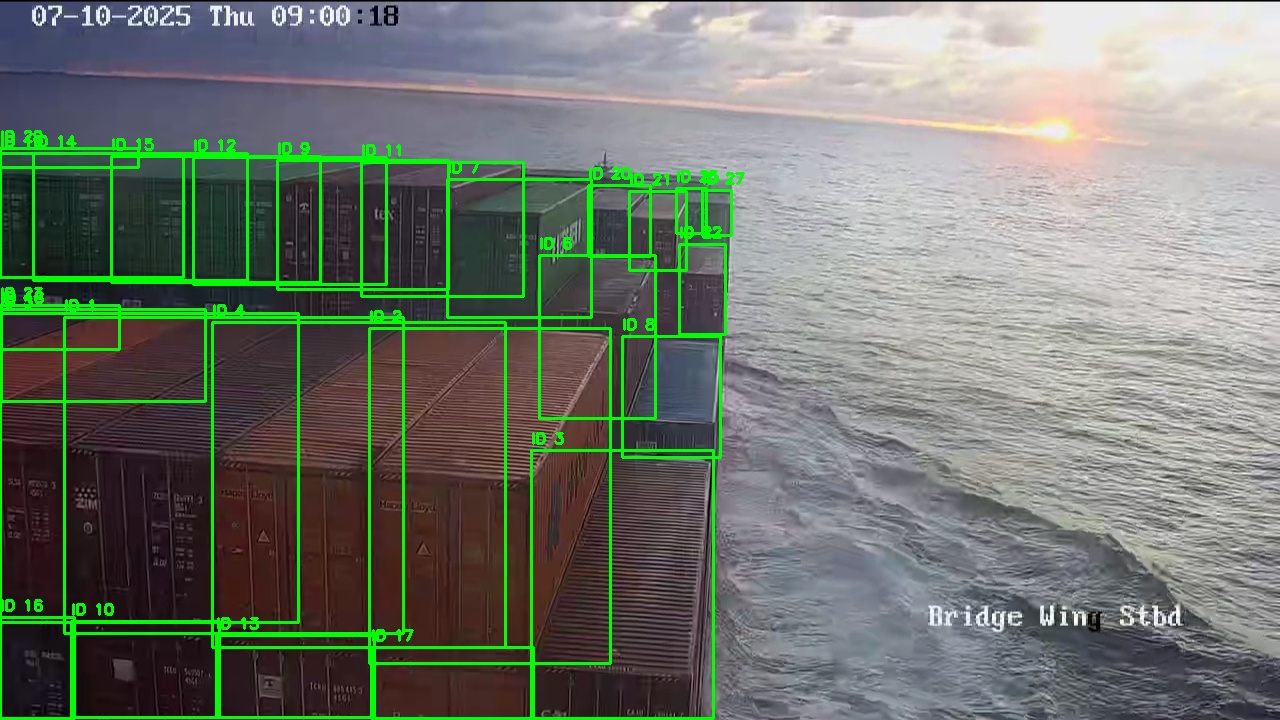}\hfill
    \caption{}
  \end{subfigure}
  \begin{subfigure}{0.18\linewidth}
    \includegraphics[width= \linewidth]{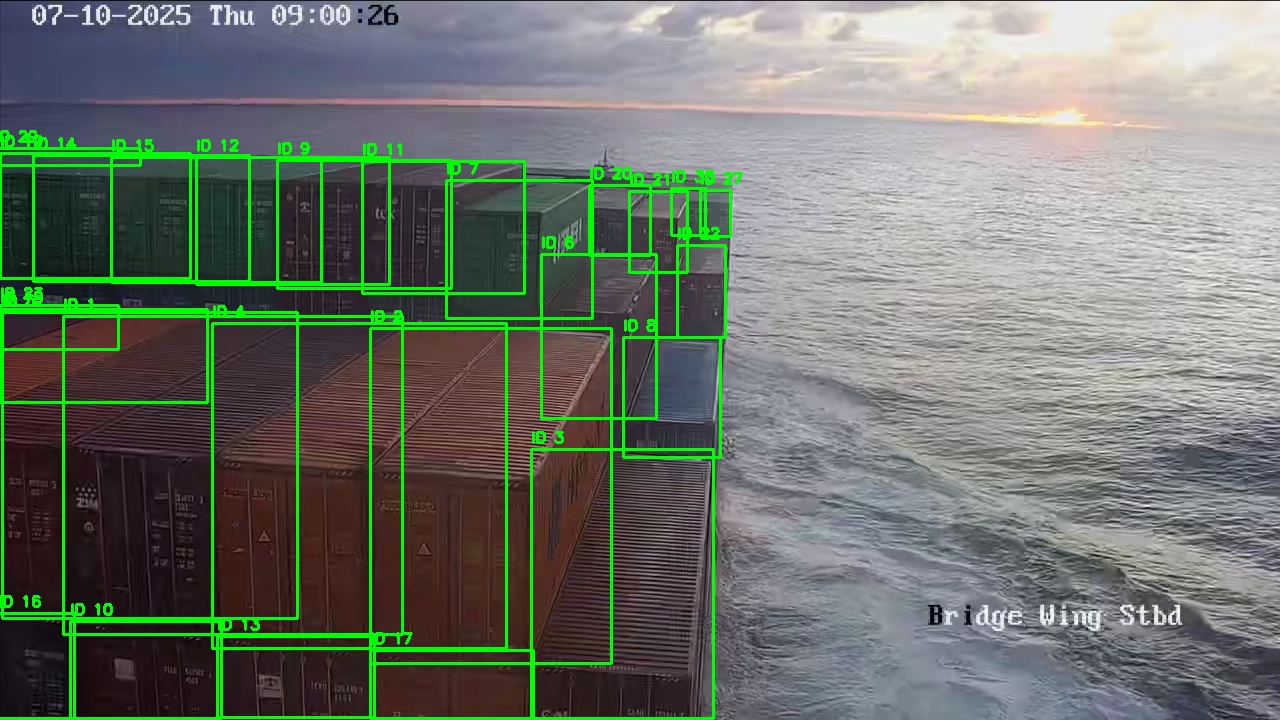}\hfill
    \caption{}
  \end{subfigure}
  \begin{subfigure}{0.18\linewidth}
    \includegraphics[width= \linewidth]{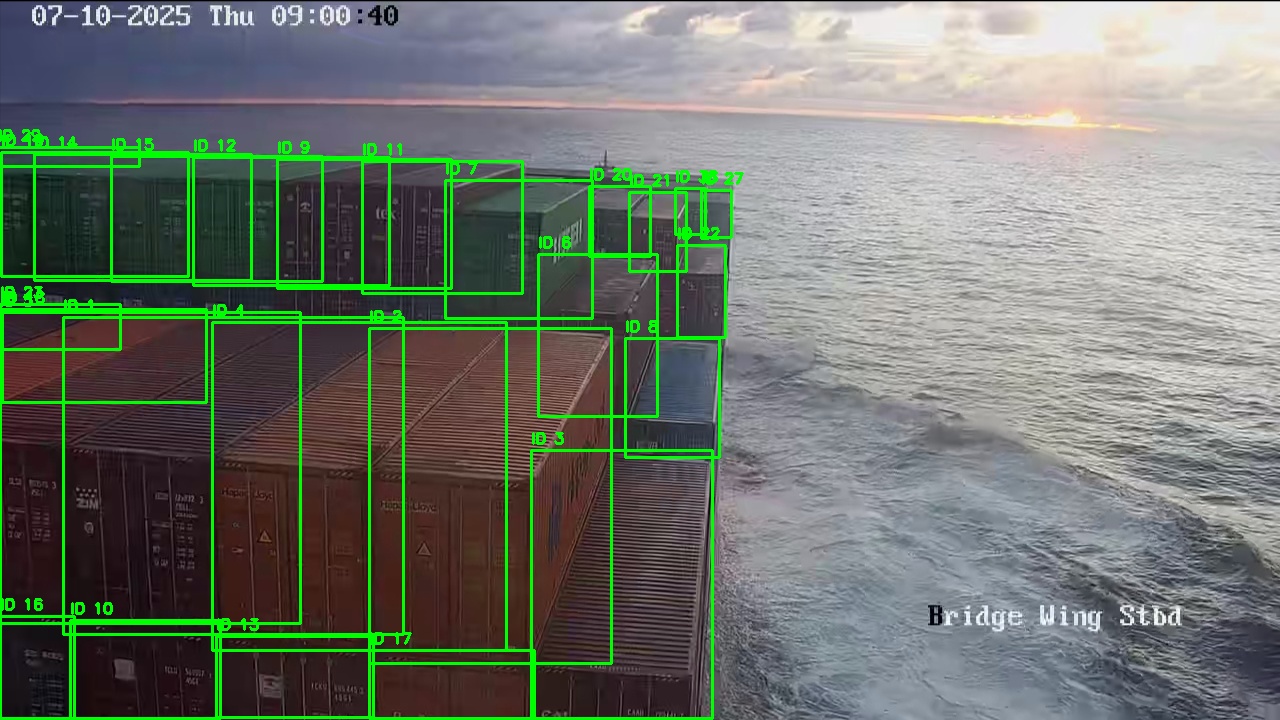}\hfill
    \caption{}
  \end{subfigure}
  \begin{subfigure}{0.18\linewidth}
    \includegraphics[width= \linewidth]{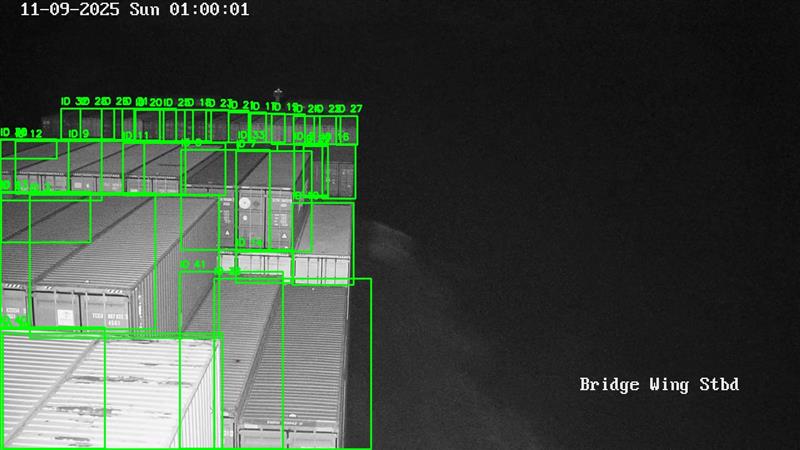}\hfill
    \caption{}
  \end{subfigure}
  \begin{subfigure}{0.18\linewidth}
    \includegraphics[width= \linewidth]{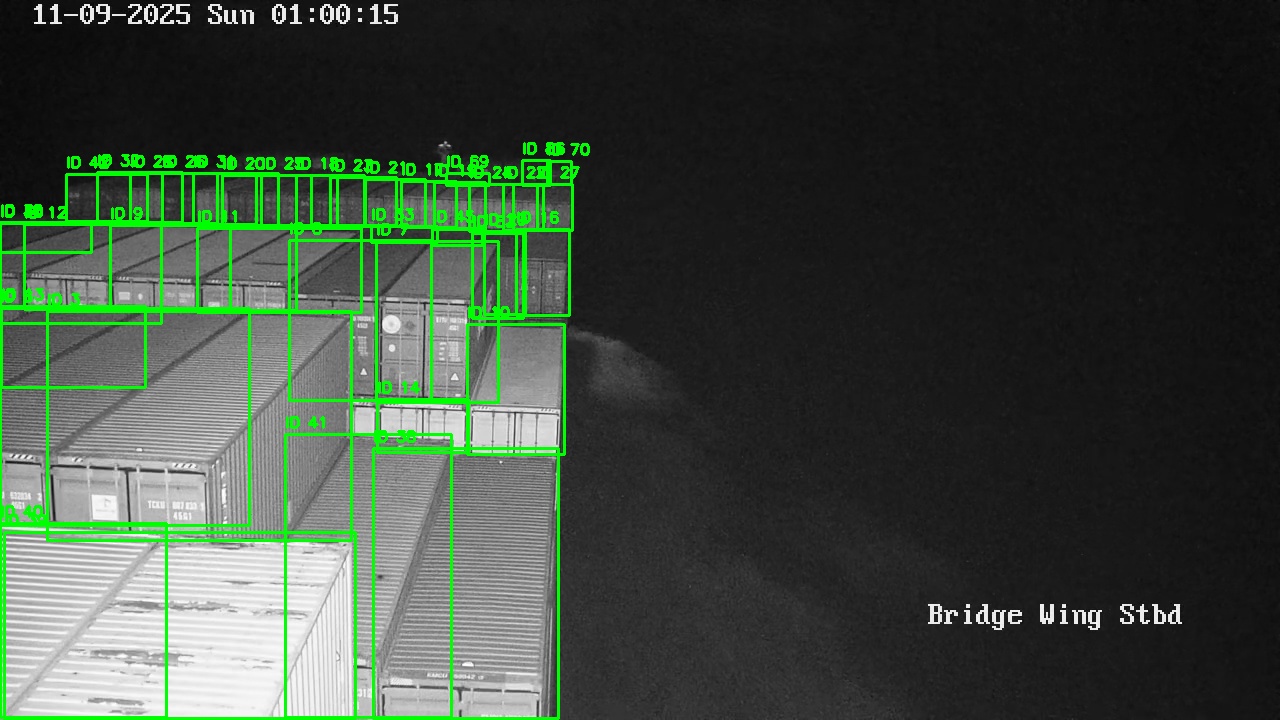}\hfill
    \caption{}
  \end{subfigure}
  \begin{subfigure}{0.18\linewidth}
    \includegraphics[width= \linewidth]{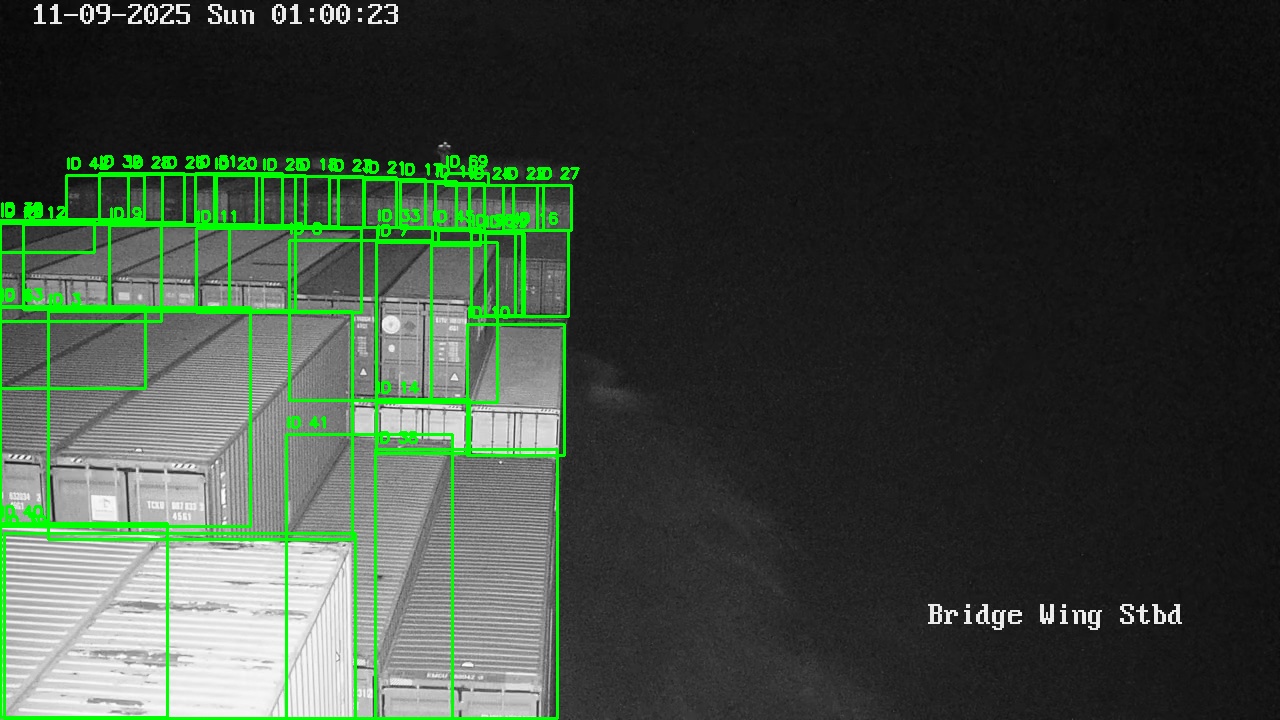}\hfill
    \caption{}
  \end{subfigure}
  \begin{subfigure}{0.18\linewidth}
    \includegraphics[width= \linewidth]{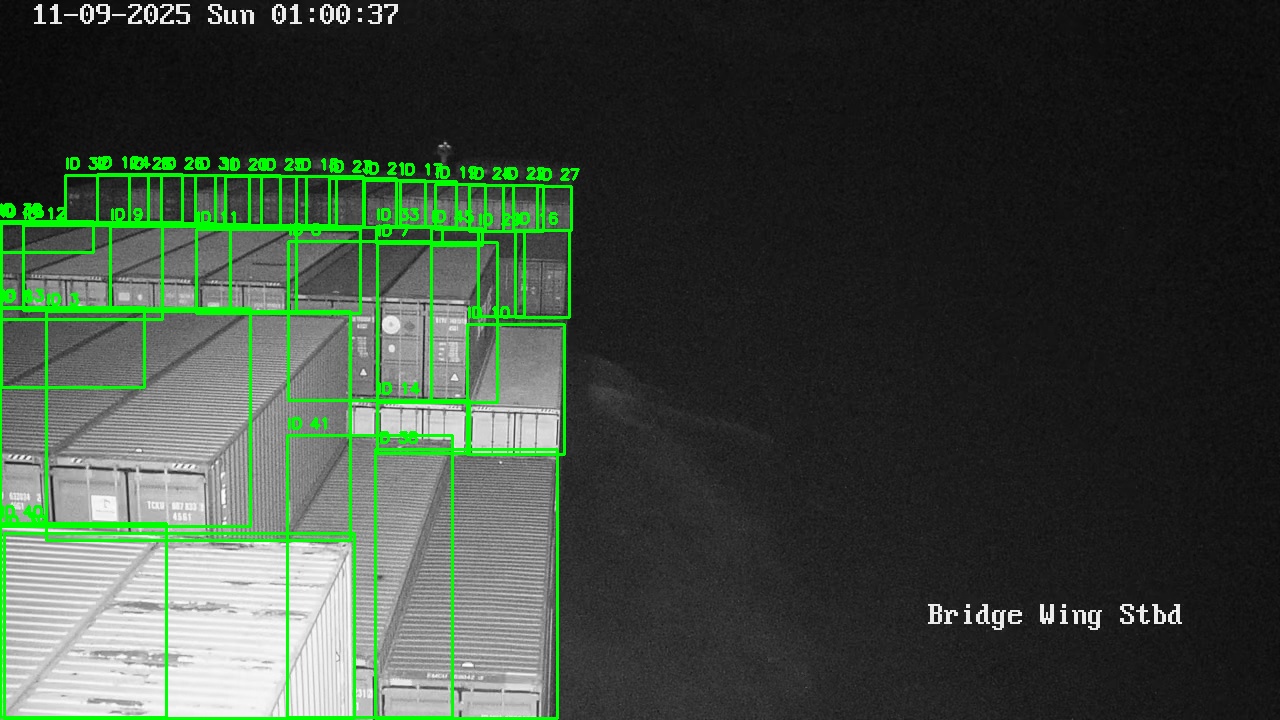}\hfill
    \caption{}
  \end{subfigure}
  \begin{subfigure}{0.18\linewidth}
    \includegraphics[width= \linewidth]{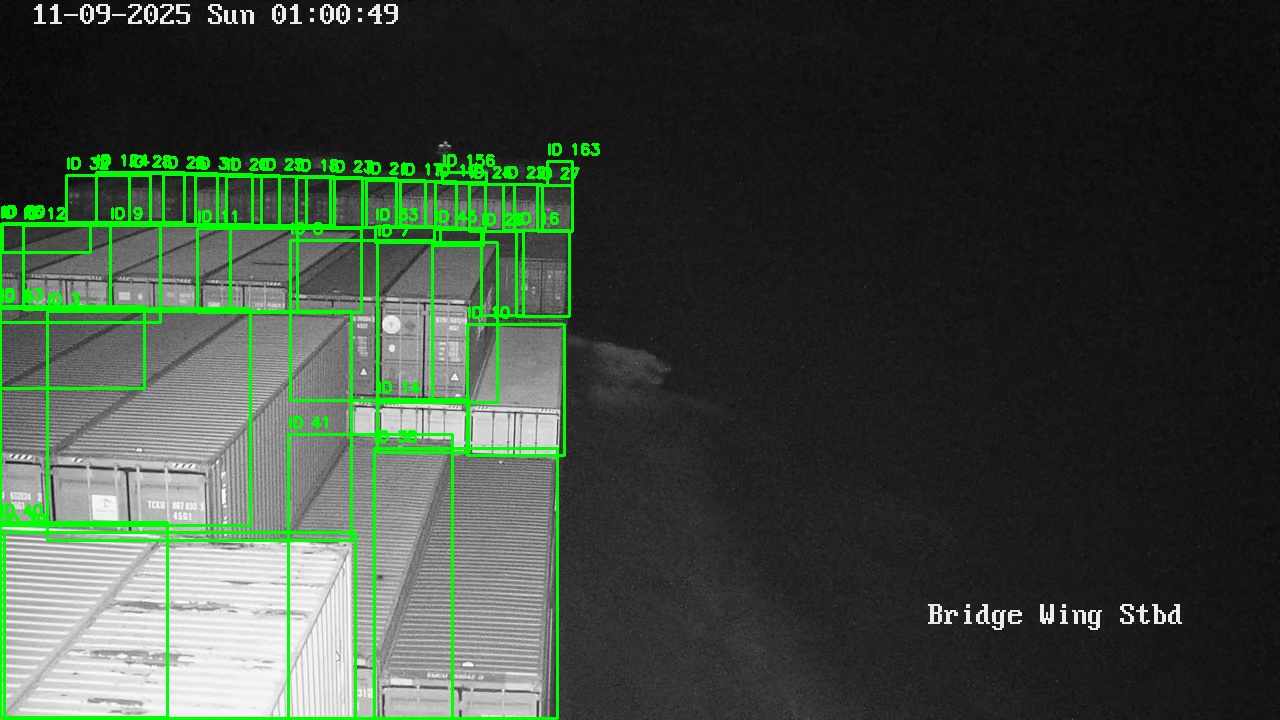}\hfill
    \caption{}
  \end{subfigure}
  \begin{subfigure}{0.18\linewidth}
    \includegraphics[width= \linewidth]{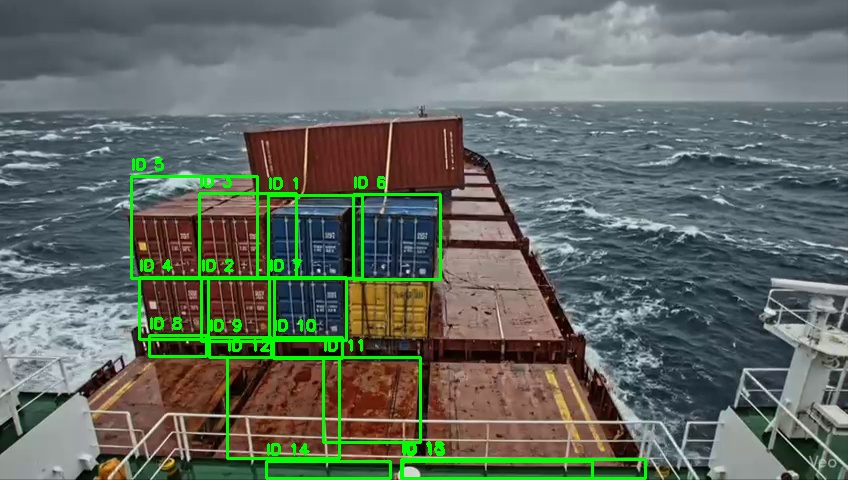}\hfill
    \caption{}
  \end{subfigure}
  \begin{subfigure}{0.18\linewidth}
    \includegraphics[width= \linewidth]{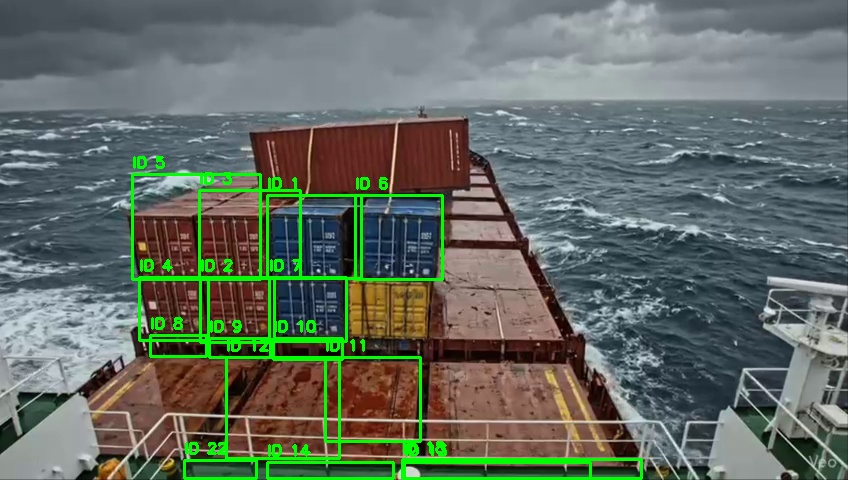}\hfill
    \caption{}
  \end{subfigure}
  \begin{subfigure}{0.18\linewidth}
    \includegraphics[width= \linewidth]{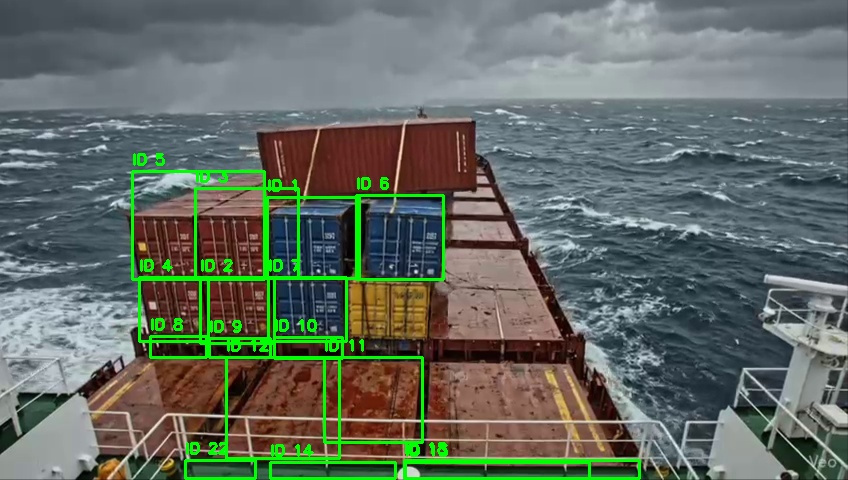}\hfill
    \caption{}
  \end{subfigure}
  \begin{subfigure}{0.18\linewidth}
    \includegraphics[width= \linewidth]{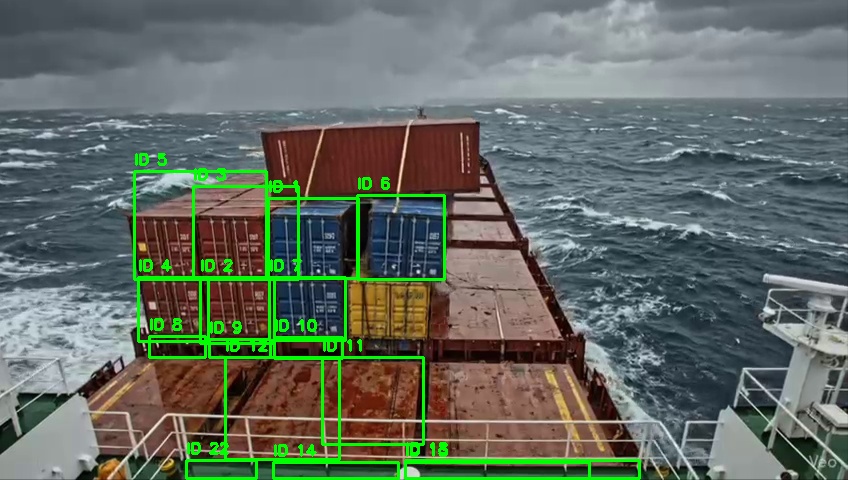}\hfill
    \caption{}
  \end{subfigure}
  \begin{subfigure}{0.18\linewidth}
    \includegraphics[width= \linewidth]{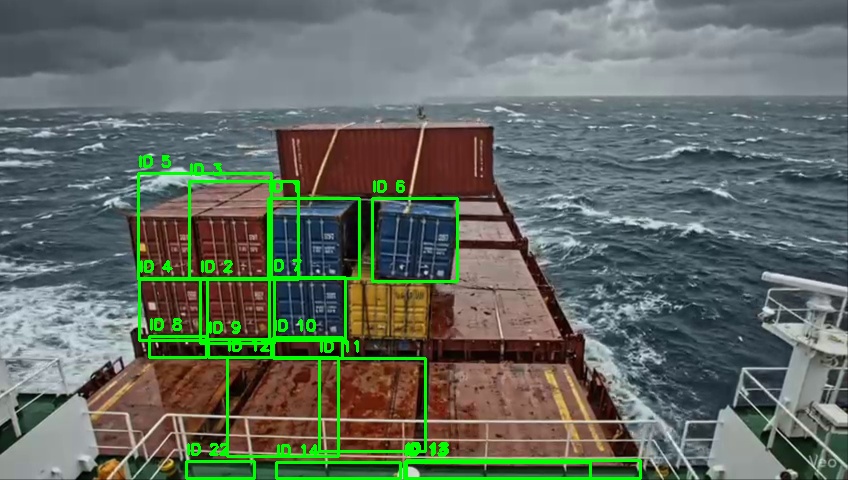}\hfill
    \caption{}
  \end{subfigure}
  \caption{Temporal tracking samples from the test dataset, each container is annotated with a bounding box and persistent tracking ID, (a:e) tracked containers in normal lighting, (f:j) tracked containers in bright lighting, (k:o) tracked in night conditions, all from collected test dataset (Sec. \ref{segmentation_model_training_dataset}), (p:t) tracked containers from simulated test video (Sec. \ref{simulated_test_dataset})}
  \label{fig:tracking_samples}
\end{figure}

Fig. \ref{fig:tracking_samples} shows that the tracking framework consistently maintains container identities across frames, even in visually challenging conditions where contrast is low and inter-frame appearance changes are minimal. Stable tracking IDs are preserved across consecutive frames, indicating reliable data association despite the dense spatial arrangement of containers and frequent partial occlusions within container stacks.

\subsection{Relative Container Movement Results}

\begin{figure}
  \centering
  \begin{subfigure}{0.22\linewidth}
    \includegraphics[width= \linewidth]{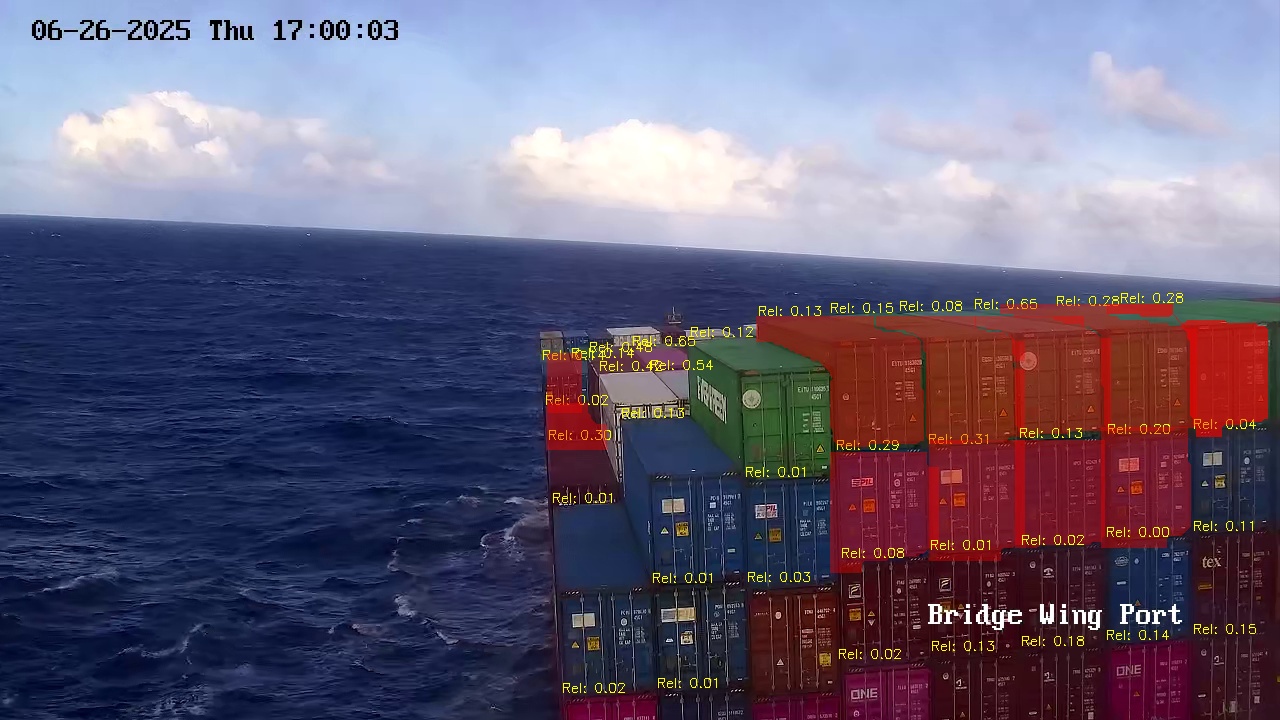}\hfill
    \caption{}
  \end{subfigure}
  \begin{subfigure}{0.22\linewidth}
    \includegraphics[width= \linewidth]{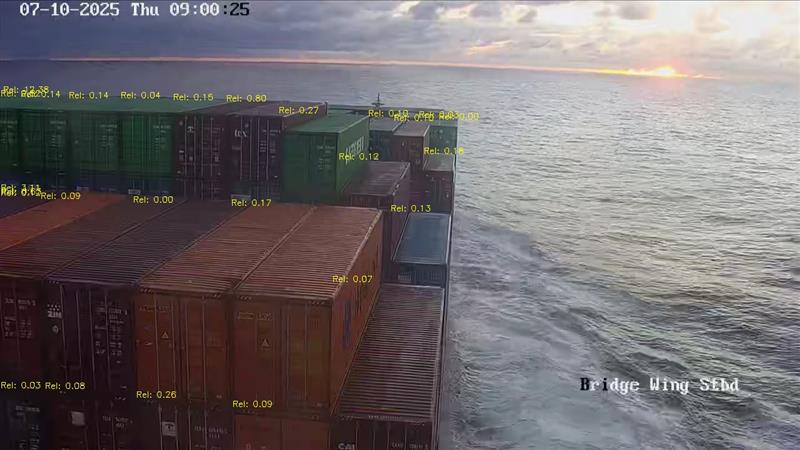}\hfill
    \caption{}
  \end{subfigure}
  \begin{subfigure}{0.22\linewidth}
    \includegraphics[width= \linewidth]{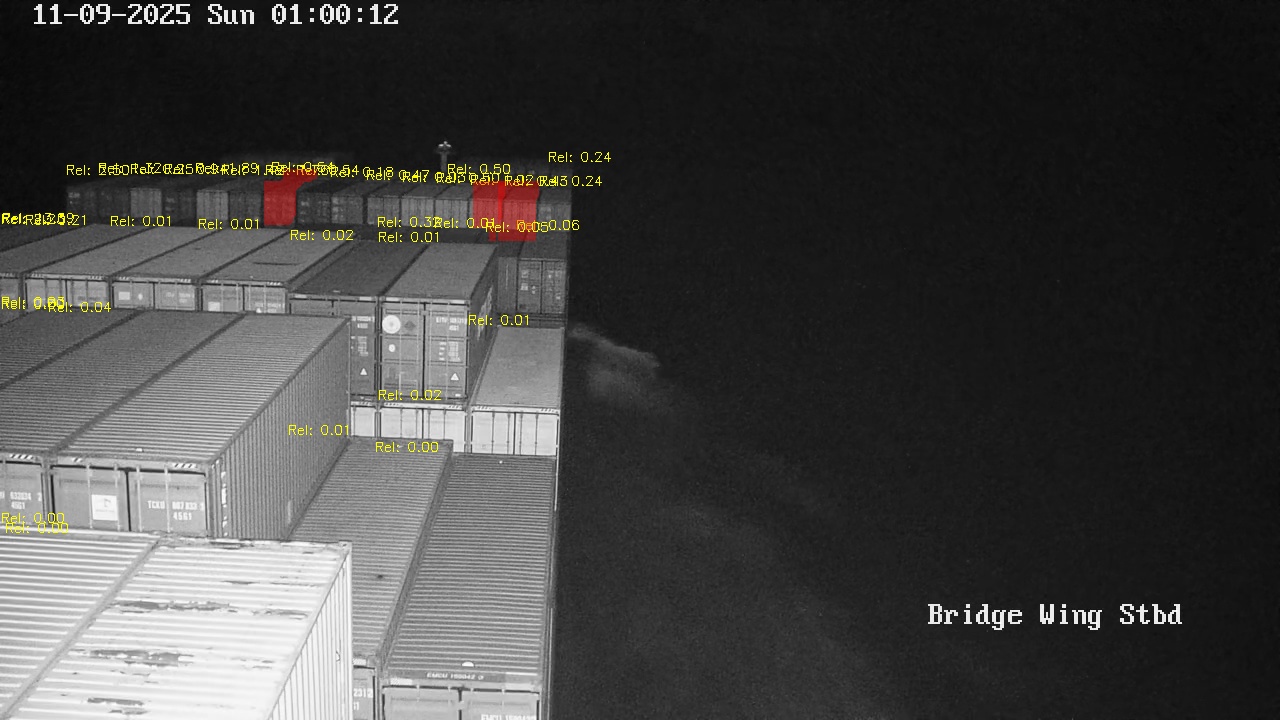}\hfill
    \caption{}
  \end{subfigure}
  \begin{subfigure}{0.22\linewidth}
    \includegraphics[width= \linewidth]{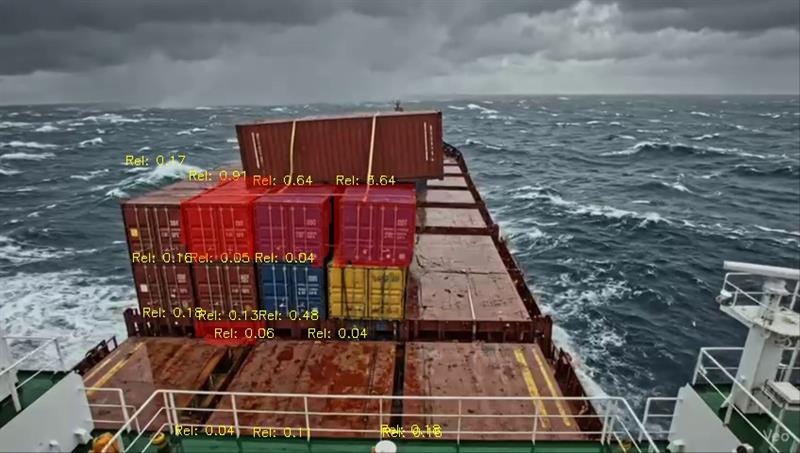}\hfill
    \caption{}
  \end{subfigure}
  \caption{Sample frames illustrating relative container motion, where per-container movement is annotated and containers exhibiting persistent instability over time are highlighted in red }
  \label{fig:movement_samples}
\end{figure}

Fig. \ref{fig:movement_samples} presents representative frames illustrating the relative motion analysis performed on individual containers after global affine motion compensation and common container motion removal. For each detected container, relative horizontal motion is estimated over time and aggregated along temporally consistent tracks.  

\begin{figure}
  \centering
  \begin{subfigure}{0.24\linewidth}
    \includegraphics[width= \linewidth]{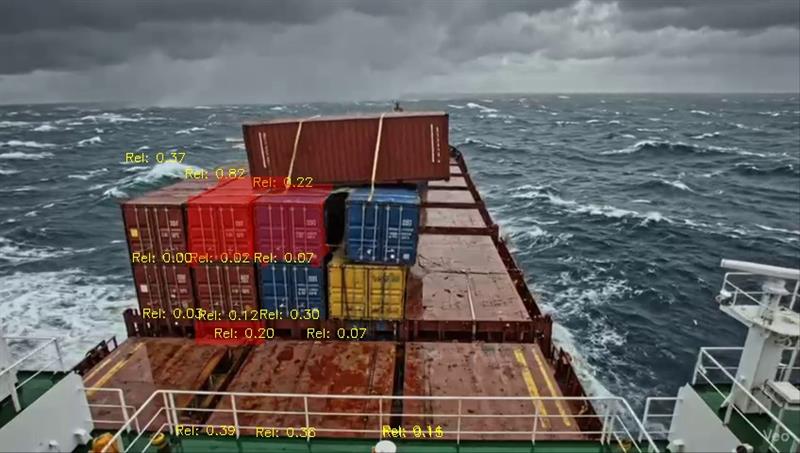}\hfill
    \caption{t=0.2 sec}
    \label{fig:early_movement_samples_1}
  \end{subfigure}
  \begin{subfigure}{0.24\linewidth}
    \includegraphics[width= \linewidth]{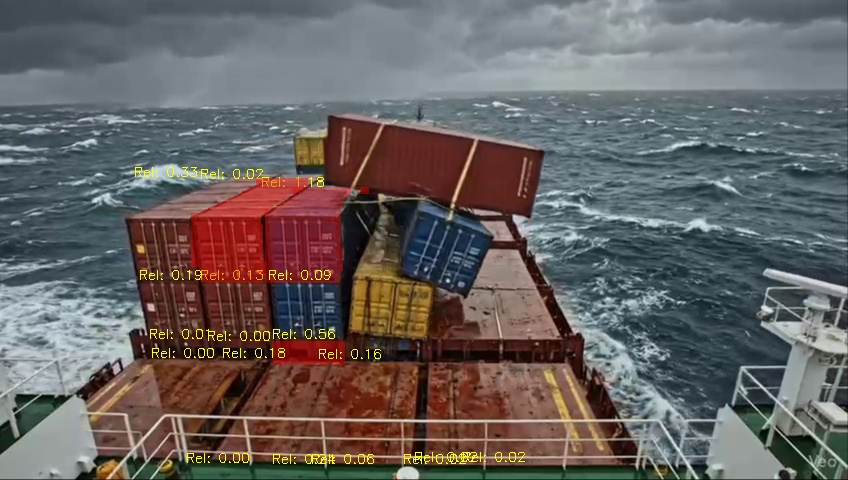}\hfill
    \caption{t=0.3 sec}
    \label{fig:early_movement_samples_2}
  \end{subfigure}
  \begin{subfigure}{0.24\linewidth}
    \includegraphics[width= \linewidth]{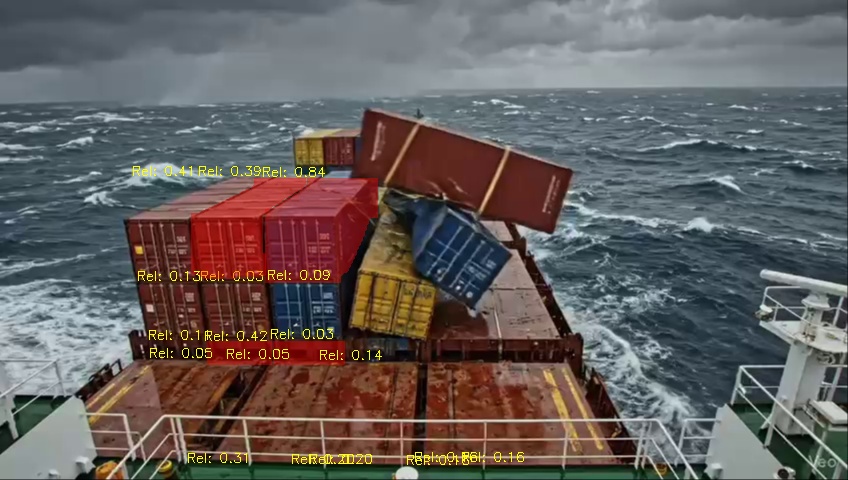}\hfill
    \caption{t=0.5 sec}
    \label{fig:early_movement_samples_3}
  \end{subfigure}
  \begin{subfigure}{0.24\linewidth}
    \includegraphics[width= \linewidth]{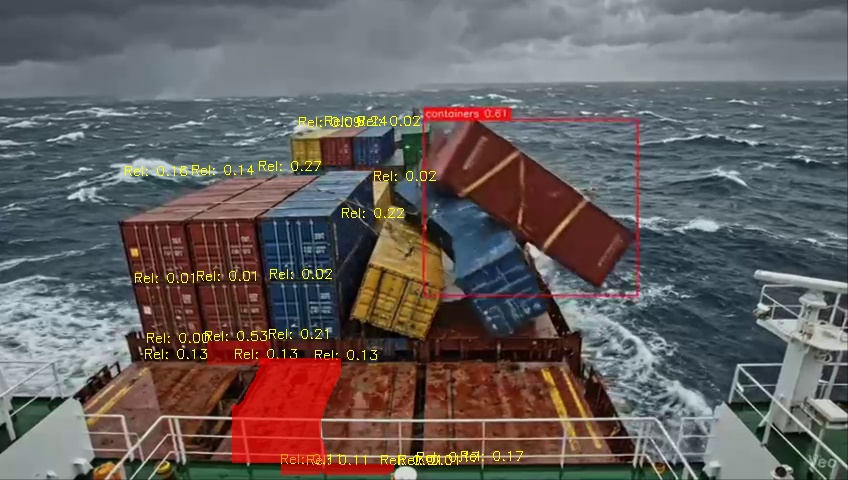}\hfill
    \caption{t=1 sec benchmark}
    \label{fig:early_movement_samples_4}
  \end{subfigure}
  \caption{Our proposed method identifies early container instability at earlier timestamps than the benchmark test dataset (Sec. \ref{simulated_test_dataset})}
  \label{fig:early_movement_samples}
\end{figure}

To benchmark the proposed approach against existing solutions, we evaluated our model on a simulated container loss video publicly released by Eyesea and EVI Safety Technologies (Sec. \ref{simulated_test_dataset}). In this video, the reference tool detects container loss approximately 1.0 seconds after the onset of container separation, corresponding to the point at which containers have already fallen and exhibit large, visually obvious displacement (Fig. \ref{fig:early_movement_samples_4}). In contrast, our method identifies container instability at earlier time instances, with observable relative motion detected at approximately 0.2 seconds (Fig. \ref{fig:early_movement_samples_1}), 0.3 seconds (Fig. \ref{fig:early_movement_samples_2}), and 0.5 seconds (Fig. \ref{fig:early_movement_samples_3}) before the detection by the commercial system. At these earlier timestamps, containers exhibit subtle but persistent instability, which is not captured by bounding-box–based detection but is revealed through mask-level optical flow analysis and temporal motion aggregation. This demonstrates the ability of the proposed approach to provide earlier warning of container destabilisation, rather than reacting only after loss has occurred.

It is important to note that the evaluated video is synthetic in nature and exhibits unrealistically rapid container motion, with a fall sequence significantly faster than what would be expected under real maritime operating conditions. While this limits the quantitative interpretation of absolute timing differences, the results nevertheless illustrate the qualitative advantage of detecting pre-failure instability rather than post-failure events. Additionally, the simulated video includes a container positioned horizontally, which does not reflect realistic stowage configurations and was not present in the training dataset. As a result, this horizontally oriented container is not detected by our segmentation model. The early instability detection results reported here are derived from containers with realistic orientations that are consistent with operational conditions.

Tab. \ref{final_results} summarises the overall movement detection performance of the proposed framework on the test dataset under varying durations of observable instability and lighting conditions. The evaluation includes both videos with visible container instability and control cases with no visible movement, enabling assessment of both detection sensitivity and false alarm behavior. For videos exhibiting observable instability, the proposed method achieves perfect detection across all evaluated conditions. Short-duration instability events (0–10 s), which are particularly challenging due to limited temporal evidence, are detected reliably under both daytime (8/8) and nighttime (25/25) conditions. Similarly, all videos with longer instability durations (10–30 s and $> 30$ s) are correctly identified, indicating that the accumulation of relative motion over time produces a strong and consistent instability signal. For videos with no visible container movement, a small number of false detections are observed: 6 out of 30 daytime samples and 5 out of 32 nighttime samples. These false positives are primarily attributed to subtle apparent motions introduced by illumination changes, sea-induced structural vibrations, and minor segmentation boundary fluctuations, which can momentarily manifest as low-magnitude relative motion. Importantly, these detections typically lack temporal persistence and can be further mitigated through stricter temporal thresholds. Overall, the results demonstrate that the proposed framework is highly sensitive to genuine container instability while maintaining a low false alarm rate.

\begin{table}[]
\centering
\caption{Test Dataset Overall Movement Detection}
\resizebox{0.7\linewidth}{!}{%
\begin{tabular}{l|l|l|l}
\textbf{\begin{tabular}[c]{@{}l@{}}Duration of obvious instability\\ (video length = 60s)\end{tabular}} & \textbf{\begin{tabular}[c]{@{}l@{}}Lighting \\ conditions\end{tabular}} & \textbf{\begin{tabular}[c]{@{}l@{}}Number of\\ samples\end{tabular}} & \textbf{\begin{tabular}[c]{@{}l@{}}Number of samples\\ detected having unstable \\ containers by model\end{tabular}} \\ \hline
No visible movement & day & 30 & 6  \\
No visible movement & night & 32 & 5  \\
0-10 s & day & 8 & 8 \\
0-10 s & night & 25 & 25 \\
10-30 s & night & 2 & 2 \\
\textgreater{}30 s & day & 2 & 2 \\
\textgreater{}30 s & night & 1 & 1
\end{tabular}%
}
\label{final_results}
\end{table}

\section{Conclusion}
This work presents a vision-based framework for early detection of container instability using ship-mounted video feeds, with the objective of identifying unsafe container motion prior to loss events. Motivated by the limitations of existing systems that primarily detect containers only after they have fallen, the proposed approach focuses on capturing subtle, relative movements of individual containers over time as precursors to instability.

A key contribution of this study is the use of instance-level container segmentation combined with temporal tracking and motion analysis. By employing a fine-tuned YOLO11 segmentation model, individual containers are precisely delineated, enabling reliable association across frames and accurate tracking. Global affine motion compensation is applied to remove common motion shared by all containers, isolating container-specific residual motion that is indicative of lashing slackness or structural instability. The integration of dense optical flow with DeepSORT-based temporal tracking allows both fine-grained motion estimation and consistent identity preservation, which is essential for persistent instability detection.

Experimental results on approximately 100 real-world vessel videos demonstrate that the proposed method is robust across diverse operational conditions, including day and night scenarios, varying visibility, and different sea states. The system consistently detects container instability across all videos exhibiting observable movement, including short-duration events lasting less than 10 seconds. While a small number of false alarms were observed in videos with no visible movement, these were limited in frequency and primarily attributable to environmental effects such as illumination changes and minor structural vibrations.

Evaluation on a publicly available simulated video further highlights the advantage of the proposed approach over commercial tools, demonstrating earlier detection of unstable containers before loss occurs. Although certain unrealistic configurations in the simulated data, the model still successfully identified early instability cues for realistic container orientations.

Overall, this study demonstrates the feasibility and effectiveness of vision-based, instance-level motion analysis for early container instability detection. The proposed framework provides a practical foundation for proactive maritime safety systems, with potential for integration into onboard monitoring solutions to support timely alerts and risk mitigation. 

\bibliography{references}

\end{document}